\documentclass{article}

\usepackage[final]{styles/corl_2026} 
\makeatletter
\renewcommand{\@conferencelocation}{Austin, Texas, USA}
\makeatother

\usepackage[utf8]{inputenc} 
\usepackage[T1]{fontenc}    
\usepackage{pdfcomment}
\usepackage{hyperref}       
\usepackage{fontawesome5}   
\usepackage{url}            
\usepackage{booktabs}       
\usepackage{amsfonts}       
\usepackage{nicefrac}       
\usepackage{microtype}      
\usepackage{xcolor}         
\usepackage{threeparttable}

\usepackage{float}
\usepackage[normalem]{ulem}
\usepackage{array}
\usepackage{graphicx}
\usepackage{subcaption}
\usepackage{wrapfig}
\usepackage{comment}
\usepackage{placeins} 
\usepackage[export]{adjustbox} 
\usepackage{titletoc}
\usepackage{multirow}

\usepackage{microtype}
\usepackage{makecell}
\usepackage{pifont}
\usepackage{pgfplots}
\pgfplotsset{compat=1.18}

\usepackage{caption}
\usepackage{amsmath}
\usepackage{mathtools}
\usepackage{amsthm}

\usepackage{enumitem}
\usepackage{bm}
\usepackage{tcolorbox}
\usepackage{needspace}
\usepackage{afterpage}

\usetikzlibrary{fit,matrix}

\newtcolorbox{conclusionbox}{
  colback=blue!5,        
  colframe=blue!75!black, 
  coltitle=black,        
  fonttitle=\bfseries,   
  boxrule=1pt,           
  arc=1mm,               
  left=2mm,              
  right=2mm,             
  top=1mm,               
  bottom=1mm,            
}

\usepackage[capitalize,noabbrev]{cleveref}
\crefname{equation}{Eq.}{Eqs.}
\Crefname{equation}{Eq.}{Eqs.}
\crefname{figure}{Fig.}{Figs.}
\Crefname{figure}{Fig.}{Figs.}
\crefname{appendix}{appendix}{appendices}
\Crefname{appendix}{appendix}{appendices}

\theoremstyle{plain}

\theoremstyle{definition}

\theoremstyle{remark}

\usepackage[textsize=tiny]{todonotes}

\usepackage{amsmath,amssymb} 

\usepackage{tikz}
\usetikzlibrary{arrows.meta,positioning,calc,decorations.pathmorphing}

\title{Hierarchical Planning with Latent World Models }
\author{
  Wancong Zhang$^{1,2,*}$ \quad Basile Terver$^1$ \quad Artem Zholus$^{1,3}$ \quad Soham Chitnis$^2$ \\
  \bfseries Harsh Sutaria$^2$ \quad Mido Assran$^1$ \quad Randall Balestriero$^{1,4}$ \quad Amir Bar$^1$ \\
  \bfseries Yann LeCun$^{2,\dagger}$ \quad Nicolas Ballas$^{1,\dagger}$ \\[4pt]
  {\normalfont $^1$FAIR at Meta \quad $^2$New York University} \\
  {\normalfont $^3$Mila - Québec AI Institute \quad $^4$Brown University} \\
  {\normalfont\small $^*$Work done at Meta \quad $^\dagger$Joint last author} \\[8pt]
  {\normalfont\small Correspondence: \href{mailto:wz1232@nyu.edu}{\texttt{wz1232@nyu.edu}} \qquad
  \href{https://kevinghst.github.io/HWM/}{\faGlobe\ Website} \qquad
  \href{https://github.com/kevinghst/HWM_PLDM}{\faGithub\ Code}}
}

\hypersetup{
  pdftitle={Hierarchical Planning with Latent World Models},
  pdfauthor={Wancong Zhang, Basile Terver, Artem Zholus, Soham Chitnis, Harsh Sutaria, Mido Assran, Randall Balestriero, Amir Bar, Yann LeCun, Nicolas Ballas}
}

\begin{document}
\maketitle
\vspace{\dimexpr-1em-8pt\relax} 


\begin{abstract}
World models are a promising path to zero-shot embodied control through planning. However, existing world model planners struggle on long-horizon, multi-stage tasks: prediction errors compound and naive search is exponential in the planning horizon. Hierarchy mitigates both by decomposing tasks into shorter, tractable subproblems; yet prior hierarchical approaches either amortize control into task-specific policies (hierarchical RL) or assume low-dimensional states and known dynamics (classical hierarchical MPC). We present Hierarchical Planning with Latent World Models (HWM), an architecture and planning paradigm for hierarchical model predictive control directly on visual world models trained solely via next-latent prediction. HWM learns world models at multiple temporal scales within a shared latent space, so predictions from the long horizon model serve as subgoals for the short horizon model via latent matching—without task-specific rewards or hierarchical policies. To keep long-horizon search tractable, HWM learns an action encoder that compresses primitive action chunks into latent macro-actions. On real-world Franka manipulation, HWM solves pick-\&-place from a single goal image at 70\% success vs. 0\% for single-level planning.
Across simulated push manipulation and maze navigation, HWM consistently improves performance on long-horizon tasks while requiring up to 3× less planning compute.

\end{abstract}



\begin{figure*}[t]
  \centering

  \resizebox{0.9\textwidth}{!}{%
    \input{figures/tikz_figures/hwm_plan_robot_feedback.tikz}%
  }

  \includegraphics[width=1.0\textwidth]{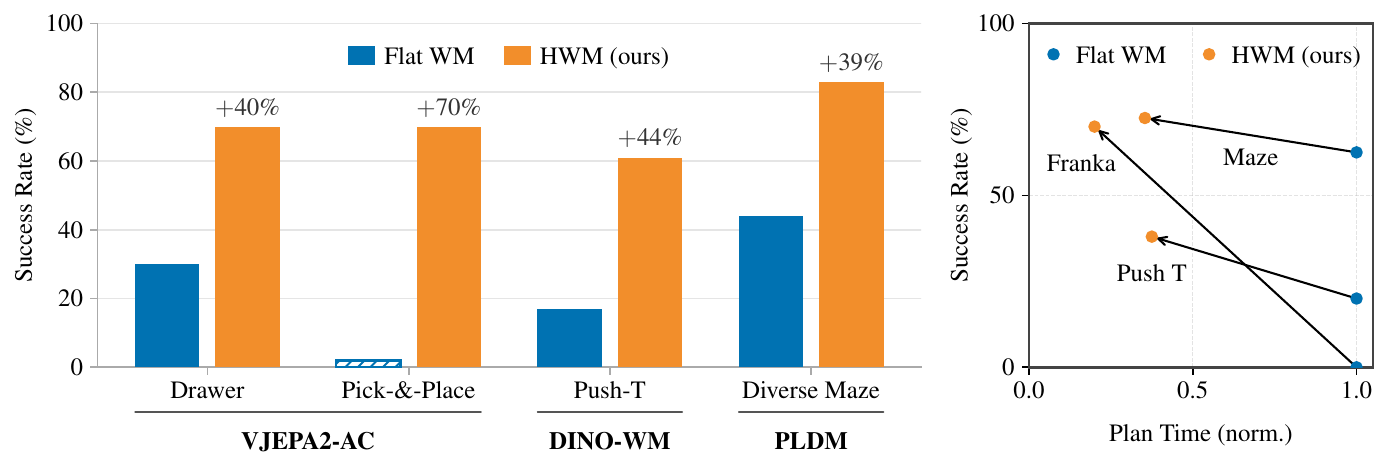}

\caption{\textbf{Top: Hierarchical planning in latent space.} A high-level planner optimizes macro-actions using a long-horizon world model to reach the goal; the first predicted latent state from the high-level rollout serves as the subgoal for a low-level planner, which optimizes primitive actions with a short-horizon world model. Both levels replan as new observations arrive, in a closed-loop MPC fashion. Solid borders denote ground-truth observations; others are decoder reconstructions shown for interpretability only. \textbf{Bottom Left:} HWM improves long-horizon planning on 3 latent world model architectures across 4 task suites. \textbf{Bottom Right:} Success rate versus normalized planning time. HWM matches or exceeds performance with up to $3\times$ lower planning time.}
  \label{fig:hierarchical_plan_plus_coverage}
  \vspace{-1em}
\end{figure*}

\section{Introduction}
\label{sec:intro}

Planning with learned world models has emerged as a powerful paradigm for embodied decision making: by simulating the consequences of candidate actions, an agent can reason about long-term outcomes without acting in the environment~\citep{sutton1981adaptive,deisenroth2011pilco,ha2018worldmodels}. Recent advances in latent-space modeling and self-supervised learning have made it possible to train such models directly from large collections of unlabeled, task-agnostic offline trajectories, enabling zero-shot planning from high-dimensional observations such as pixels~\citep{assran2025v, sobal2025learning, bar2025navigation, zhou2024dino, goswami2025world}. Yet single-level planning with these models fails in two distinct regimes. {On tasks with non-greedy objectives, such as pick-and-place, the optimal trajectory must temporarily move away from the goal. The resulting non-monotone goal-matching cost can trap local optimizers in greedy shortcuts even at modest horizons}; empirically, even simple real-world pick-and-place is out of reach for state-of-the-art world-model planners such as VJEPA2-AC \citep{assran2025v} (\cref{tab:franka}). On long-horizon tasks, prediction errors compound over autoregressive rollouts~\citep{talvitie2014model, janner2019trust} and the action search space grows exponentially with horizon~\citep{ichter2020broadly}.

Introducing temporal hierarchy into decision-making — reasoning coarsely over long horizons while retaining fine-grained control at short horizons — has long been studied as a way to mitigate both effects, but existing approaches do not transfer cleanly to zero-shot planning on visual world models. Hierarchical reinforcement learning based on options, skills, or hierarchical policies~\citep{sutton1999between, bacon2017option, hafner2022deep, park2023hiql} requires task-specific rewards or task distributions, and even modern goal-conditioned and zero-shot RL methods~\citep{park2023hiql, park2024foundation} generalize poorly beyond training~\citep{sobal2025learning}. Hierarchical world models have been explored in the RL setting~\citep{gumbsch2023learning, schiewer2024exploring, hansen2024hierarchical}, but remain coupled to policy learning over a specific task distribution. Hierarchical MPC in optimal control~\citep{fang2019dynamics, li2021planning, kogel2025safe} is task-agnostic but has been confined to low-dimensional states, hand-engineered features, or known dynamics. It remains unclear how to realize hierarchical zero-shot planning directly on visual world models.

We propose \emph{Hierarchical Planning with Latent World Models} (HWM), a framework for zero-shot hierarchical MPC over learned latent world models (Table~\ref{tab:hierarchical_comparison} contrasts HWM with prior hierarchical model-based methods). HWM trains world models at multiple temporal resolutions within a shared latent space, supervised solely by next-latent prediction, and couples them at planning time: predictions from the coarser model serve as subgoals that the finer-scale MPC matches in latent space; no hierarchical policy, skill, or goal-conditioned controller is needed. To support efficient long-horizon planning, we further introduce a learned action encoder that compresses sequences of primitive actions between waypoint states into latent macro-actions, reducing the dimensionality of the coarser-scale search. To our knowledge, HWM is the first world-model planner to demonstrate zero-shot, {real-robot manipulation under non-greedy objectives} from pixels with a single goal image, solving Franka pick-\&-place at 70\% success while single-level VJEPA2-AC planner \citep{assran2025v} achieves 0\%.


\begin{figure*}[t]
    \centering
    \resizebox{\textwidth}{!}{\usetikzlibrary{arrows.meta,positioning,calc}
\usetikzlibrary{shapes.gates.logic.US}
\usetikzlibrary{decorations.pathreplacing}
\definecolor{ZBase}{HTML}{4E79A7}     
\definecolor{ActBase}{HTML}{F7C9A9}   
\definecolor{LatBase}{HTML}{CC4C02}   
\definecolor{LossBase}{HTML}{B07AA1}
\definecolor{EncBase}{HTML}{AFC6E9}

\tikzset{
  arr/.style={-{Stealth[length=2.2mm]}, line width=0.9pt, draw=black!85},
  arrd/.style={arr, dashed, draw=black!60},
  opt/.style={->, dashed, line width=1.0pt, draw=red!80!black},
  optlab/.style={font=\bfseries\small, text=red!80!black},
  circ/.style={circle, line width=0.9pt, minimum size=9mm, inner sep=0pt},
  tok/.style={
    draw=black!60, rounded corners=2.2pt, line width=0.8pt,
    minimum width=7mm, minimum height=7mm, inner sep=1pt,
    text=black!85,
    font=\normalsize
  },
  stok/.style={tok},
  ztok/.style={tok, fill=ZBase!16},
  atok/.style={tok, fill=ActBase!30},   
  ltok/.style={tok, fill=LatBase!40},   
optvar/.style={
  draw=red!75!black,
  line width=0.95pt,
  fill=red!75!black!10,
  rounded corners=2.6pt,
  minimum size=7mm,
  inner xsep=1.0pt,
  inner ysep=1.0pt,
  outer sep=0pt,
  font=\normalsize,
  text=red!75!black,
},
encgate/.style={
  and gate US,
  logic gate inputs=nn,
  rotate=90,
  transform shape,
  xscale=1.0,          
  yscale=1.3,          
  draw=ZBase!60!black, fill=EncBase!35,
  line width=0.9pt,
  minimum width=10mm,  
  minimum height=3mm,  
  inner xsep=5pt, inner ysep=2pt
},
  enctext/.style={font=\bfseries, text=black!90},
  block/.style={
    and gate US,
    logic gate inputs=nn,  
    draw=black!80, fill=black!3,
    line width=0.9pt,
    minimum width=8mm, minimum height=8mm,
    font=\bfseries, text=black!95,
    inner xsep=5pt, inner ysep=2pt
  },
loss/.style={
  draw=red!85!black,       
  dotted,
  line width=1.1pt,        
  fill=red!85!black!8,     
  rounded corners=3.2pt,
  minimum size=9mm,
  inner xsep=1.2pt, inner ysep=1.0pt,
  font=\bfseries\small,
  text=red!85!black
},
optptr/.style={opt, dotted, line width=1.0pt} 
}

\ifhmode\loop\ifdim\lastskip>0pt\unskip\repeat\fi
\begin{tikzpicture}[font=\small]

\def\xS{0.0}

\def\xPtop{3.6}      
\def\xPbot{1.8}      

\def\xZt{6.0}        
\def\xPmidTop{8.5}   
\def\xPmidBot{4.0}   

\def\xPendA{11.0}    
\def\xPendB{12.6}    
\def\xZH{15.1}

\def\xZh{6.0}

\def\xGoal{\xZH}

\def\yTop{5.6}

\def\yZfixed{2.4}   
\def\dzEZ{1.6}      
\def\dzSE{1.6}      

\def\yMid{\yZfixed}

\def\yL{\yTop-1.35}

\node[ztok] (z1) at (\xS,\yZfixed) {$z_1$};

\node[encgate] (E1) at (\xS,{\yZfixed-\dzEZ}) {};
\node[enctext] at (E1.center) {$E$};

\node[stok] (s1) at (\xS,{\yZfixed-\dzEZ-\dzSE}) {$s_1$};

\coordinate (E1in)  at ($(E1.input 1)!0.5!(E1.input 2)$);
\coordinate (E1out) at (E1.output);
\draw[arr] (s1.north) -- (E1in);
\draw[arr] (E1out) -- (z1.south);

\node[block] (P2a) at (\xPtop,\yTop) {$F^{(2)}$};
\node[optvar] (l1)  at (\xPtop,\yL) {$l_{t_1}$};
\node[ztok]  (zt2) at (\xZt,\yTop) {$\hat z_{t_2}$};

\draw[arr] (z1.north) |- (P2a.west);
\draw[arr] (l1) -- (P2a.south);
\draw[arr] (P2a.east) -- (zt2.west);

\node[block] (P2b) at (\xPmidTop,\yTop) {$F^{(2)}$};

\node[optvar] (l2)  at (\xPmidTop,\yL) {$l_{t_2}$};
\draw[arr] (l2) -- (P2b.south);

\node[block] (P2c) at (\xPendB,\yTop) {$F^{(2)}$};
\node[optvar] (lH)  at (\xPendB,\yL) {$l_{t_H}$};
\node[ztok]  (zH)  at (\xZH,\yTop) {$\hat z_{t_{H+1}}$};

\draw[arr]  (zt2.east) -- (P2b.west);

\draw[arrd,-] (P2b.east) -- (\xPendA,\yTop);
\draw[arrd]   (\xPendA,\yTop) -- (P2c.west);

\node[text=black!70] at ($(l2)!0.5!(lH)$)
  {\Large $ \cdot\ \ \cdot\ \ \cdot\ \ \cdot\ \ \cdot$};

\draw[arr]  (lH) -- (P2c.south);
\draw[arr]  (P2c.east) -- (zH.west);

\node[block] (P1a) at (\xPbot,\yMid) {$F^{(1)}$};
\node[optvar] (a1)  at (\xPbot,\yMid-1.35) {$a_1$};
\node[block] (P1b) at (\xPmidBot,\yMid) {$F^{(1)}$};
\node[optvar] (ah)  at (\xPmidBot,\yMid-1.35) {$a_h$};

\node[ztok]  (zh)  at (\xZh,\yMid) {$\hat z_{h+1}$};

\draw[arr] (z1.east) -- (P1a.west);

\draw[arr]  (a1) -- (P1a.south);
\draw[arrd] (P1a.east) -- (P1b.west);
\draw[arr]  (ah) -- (P1b.south);
\draw[arr]  (P1b.east) -- (zh.west);
\node[text=black!70] at ($(a1)!0.5!(ah)$) {\Large $\dots$};
\node[ztok] (zg) at (\xGoal,\yZfixed) {$z_{\text{goal}}$};

\node[encgate] (Eg) at (\xGoal,{\yZfixed-\dzEZ}) {};
\node[enctext] at (Eg.center) {$E$};

\node[stok] (sg) at (\xGoal,{\yZfixed-\dzEZ-\dzSE}) {$s_{\text{goal}}$};

\coordinate (Egin)  at ($(Eg.input 1)!0.5!(Eg.input 2)$);
\coordinate (Egout) at (Eg.output);
\draw[arr] (sg.north) -- (Egin);
\draw[arr] (Egout) -- (zg.south);

\node[loss] (L1) at ({(\xZt+\xZh)/2},\yL) {$\mathcal{L}_{\text{low}}$};
\draw[arrd] (zt2.south) -- ($(zt2.south)+(0,-0.35)$) -| (L1.north);
\draw[arrd] (zh.north)  -- ($(zh.north)+(0,0.35)$)   -| (L1.south);

\node[loss] (L2) at (\xZH,\yL) {$\mathcal{L}_\text{high}$};
\draw[arrd] (zH.south) -- (L2.north);
\draw[arrd] (zg.north) -- (L2.south);


\node[optlab, align=center] (opt2lab) at (l2 |- 0,{\yL-2.5})
  {
High-Level Planning\\[-2pt]
to Goal\\[-2pt]
$\arg\min_{\{l\}} \mathcal{L}_{\text{high}}$

  };

\node[optlab, align=center] (opt1lab) at ({(\xPbot+\xPmidBot)/2}, {\yMid-3})
  {
Low-Level Planning\\[-2pt]
to $1^{\text{st}}$ Subgoal\\[-2pt]
$\arg\min_{\{a\}} \mathcal{L}_{\text{low}}$
  };

\draw[optptr] (opt2lab.north) to[out=90,in=-90,looseness=0.8] (l1.south);
\draw[optptr] (opt2lab.north) to[out=90,in=-90] (l2.south);
\draw[optptr] (opt2lab.north) to[out=90,in=-90] (lH.south);

\draw[optptr] (opt1lab.north) to[out=90,in=-90] (a1.south);
\draw[optptr] (opt1lab.north) to[out=90,in=-90] (ah.south);

\def\yLegend{\yMid-2.40}
\coordinate (legC) at ({0.665*\xGoal}, \yLegend);
\def\legScale{0.85}

\begin{scope}[shift={(legC)}, scale=\legScale, transform shape, font=\small]

  \draw[draw=black!35, line width=0.55pt, rounded corners=2.0pt, fill=white]
    (-6.1, 0.95) rectangle ( 4.9,-1);

  \node[font=\bfseries, text=black!85]
    at ({(-6.1+4.3)/2}, 0.55) {Model key};

  \node[block, minimum width=7.2mm, minimum height=6.2mm, font=\bfseries]
    (P1leg) at (-5.25,-0.30) {$F^{(1)}$};
  \node[anchor=west, text=black!85]
    (P1txt) at ($(P1leg.east)+(0.28,0)$) {Low-level world model};

  \node[block, minimum width=7.2mm, minimum height=6.2mm, font=\bfseries]
    (P2leg) at ($(P1txt.east)+(1.15,0)$) {$F^{(2)}$};
  \node[anchor=west, text=black!85]
    (P2txt) at ($(P2leg.east)+(0.28,0)$) {High-level world model};

\end{scope}

\end{tikzpicture}%
}
    \caption{
        \textbf{Hierarchical planning in latent space.}
        A high-level planner optimizes macro actions using a long-horizon latent world model to reach the final goal.
        The first high level latent prediction serves as a subgoal for low-level planning, where a short-horizon world model optimizes primitive actions to reach this subgoal.
}
\label{fig:hierarchical_plan}
\vspace{-4mm}
\end{figure*}
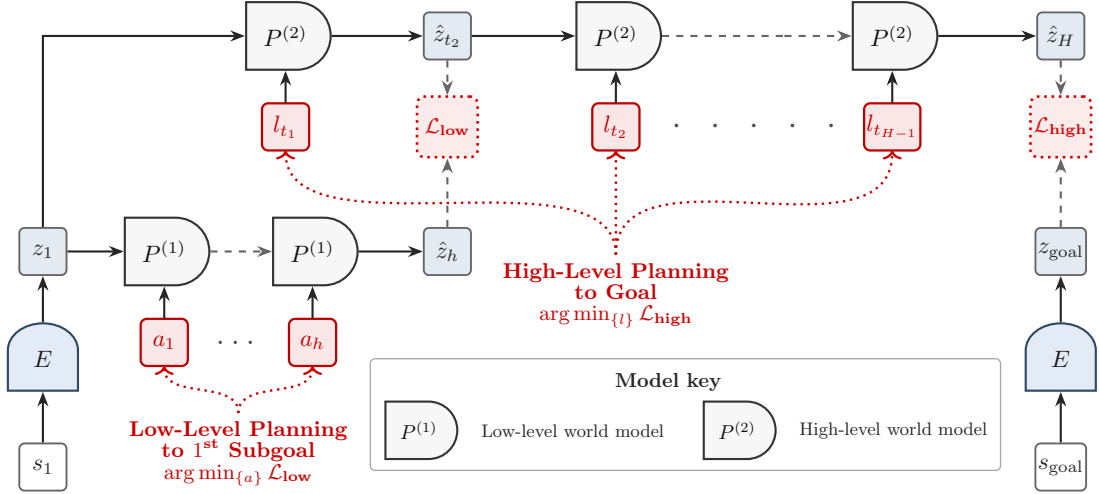

We summarize our contributions as follows:
\begin{itemize}
    \item We introduce HWM, a hierarchical MPC formulation that couples learned world models at different temporal scales via a shared latent space, enabling direct subgoal transfer across levels without hierarchical policies, skill learning, or task-specific rewards.
    \item We show that HWM unlocks a new capability for zero-shot world-model planning: {solving real-robot tasks with non-greedy objectives} from visual inputs, where success requires temporarily moving away from the goal. HWM achieves 70\% success on Franka pick-\&-place from a single goal image, compared to 0\% for VJEPA2-AC \citep{assran2025v}, and also compares favorably to several zero-shot VLA baselines under the evaluated setup.
    \item We demonstrate consistent gains across three latent world-model backbones --- VJEPA2-AC (manipulation), DINO-WM (push manipulation), and PLDM (maze navigation) --- and show that hierarchical planning improves both performance (up to $+44\%$ and $+39\%$ absolute) and efficiency (up to $3\times$ less planning compute) on long-horizon tasks.
\end{itemize}

\section{Hierarchical Planning with Latent World Models}
\label{sec:method}

\textbf{Problem Setting.}
We consider goal-conditioned control in an MDP
$\mathcal{M}=(\mathcal{S},\mathcal{A},\mu,p)$ with access to an offline dataset
$\mathcal{D}$ of state-action trajectories
$\tau=(s_1,a_1,s_2,\ldots,a_{T-1},s_T)$.
At test time, the agent receives an observation $s_t$ and a goal observation
$s_g$, both raw pixels, and must reach the goal via through
model predictive control. Given a learned world model and a planning horizon
$h$, MPC optimizes a candidate action sequence $a_{t:t+h-1}$ under a
goal-reaching objective, executes the first $k \leq h$ actions, observes the new
state, and replans. 

\textbf{Planning with Latent World Models.}
A latent world model consists of an encoder $E$ mapping observations to latent states $z_t=E(s_t)$ and a predictor $F$ forecasting future latents under candidate actions. In  latent-space MPC, the planner encodes the current and goal observations as $z_1=E(s_1)$ and $z_g=E(s_g)$, then optimizes primitive actions $a^*_{1:h}=\arg\min_{\hat a_{1:h}}\|F(\hat a_{1:h};z_1)-z_g\|_1$, where $F(\hat a_{1:h};z_1)$ is the final latent after autoregressive rollout. This policy-free formulation enables zero-shot goal-conditioned control,  but such single-level planning suffers from the two failure modes identified in \cref{sec:intro} — {getting trapped in greedy solutions under non-greedy objectives}, and compounding error and exponential action search on long-horizon tasks.  We address both by planning hierarchically over latent world models at multiple timescales: \cref{sec:multiscale_models} describes how to train these models, and \cref{sec:h_planning} describes how to use them for hierarchical planning.

\vspace{-2mm}
\subsection{Learning Multi-Timescale Latent World Models}
\label{sec:multiscale_models}

We train multiple world models operating at different time scales in a shared latent space to support hierarchical MPC. The low-level model $F^{(1)}_{\theta}$ is a standard latent world model trained to predict the next latent state (Appendix~\ref{app:l1_wm}). The high-level model $F^{(2)}_{\phi}$ predicts dynamics across temporally extended segments, conditioned on latent \emph{macro-actions} $l_t$.

\Needspace{14\baselineskip}
\begin{wrapfigure}{R}{0.45\linewidth}
\centering
\includegraphics[width=1.0\linewidth]{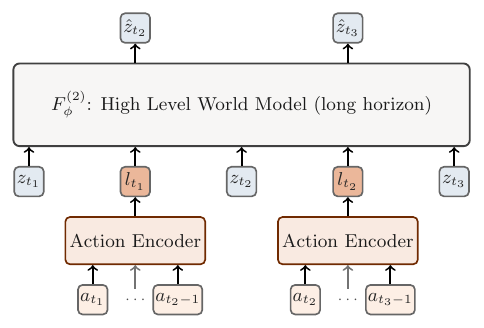}
\caption{
The high-level world model captures long-horizon dynamics using macro-actions encoded from chunks of low-level actions. It is causal and trained with a latent prediction loss.
}
\label{fig:high_lvl_wm}
\end{wrapfigure}

During training, a learned action encoder $A_{\psi}$ produces macro-actions by summarizing chunks of primitive actions. We do not require a fixed duration for these chunks, allowing each high-level transition to represent a variable-length segment of low-level execution.
This allows $F^{(2)}_{\phi}$ to overcome two limitations of single-level planning. First, fewer autoregressive rollout steps are needed to reach a goal horizon, reducing compounding error. Second, compressing primitive action chunks into a lower-dimensional latent space shrinks the high-level MPC search space, making long-horizon optimization more tractable. Learned macro-actions outperform both concatenated primitive actions and hand-crafted summaries such as net end-effector displacement (\cref{sec:ablations}).

Given a trajectory $\tau = (s_1, a_1, \ldots, s_T)$, we sample $N$ waypoint indices $1 = t_1 < \ldots < t_N \leq T$, encode waypoints as $z_{t_k} = E(s_{t_k})$ and action chunks as $l_{t_k} = A_{\psi}(a_{t_k:t_{k+1}-1})$, and feed the interleaved sequence $(l_{t_k}, z_{t_k})_{k \in [N-1]}$ to $F^{(2)}_{\phi}$ to predict the next waypoint latents $\hat{z}_{t_{k+1}} := F^{(2)}_{\phi}((l_{t_i}, z_{t_i})_{i \le k})$. We supervise with the teacher-forcing $\ell_1$ loss $\mathcal{L}_{\mathrm{tf}}(\phi, \psi) := \tfrac{1}{N-1} \sum_{k=1}^{N-1} \| \hat{z}_{t_{k+1}} - z_{t_{k+1}} \|_1$. We use this teacher-forcing objective for Franka and Push-T, and an autoregressive rollout objective for Diverse Maze (Appendix~\ref{app:l1_wm}). With both models trained, hierarchical MPC follows directly (\cref{sec:h_planning}).

\subsection{Top-Down Hierarchical Planning}
\label{sec:h_planning}

Using the trained world models from \cref{sec:multiscale_models}, we adopt a \emph{top-down} hierarchical planning strategy operating entirely in their shared latent space (Fig.~\ref{fig:hierarchical_plan}). A high-level planner optimizes macro-actions toward the goal using $F^{(2)}$, and the resulting latent predictions serve as subgoals for a low-level planner optimizing primitive actions using $F^{(1)}$, in a receding-horizon manner.

\textbf{High-Level Planning.}
Given an initial observation $s_1$ and a goal observation $s_g$, we encode them into latent space as $z_1 = E(s_1)$ and $z_g = E(s_g)$. At the high level, we plan over a sequence of $H$ macro-actions $\hat{l}_{1:H}$ by minimizing a goal-conditioned energy function in latent space:
\[
\mathcal{E}_2(\hat{l}_{1:H}; z_1, z_g)
\;\triangleq\;
\left\| z_g - F^{(2)}(\hat{l}_{1:H}; z_1) \right\|_1
\]
Here, $F^{(2)}(\hat{l}_{1:H}; z_1)$ denotes the final latent state obtained by autoregressively unrolling the world model $F^{(2)}$ starting from $z_1$ with macro-actions $\hat{l}_{1:H}$. The optimal macro-action sequence is given by:
$
l^*_{1:H}
=
\arg\min_{\hat{l}_{1:H}}
\mathcal{E}_2(\hat{l}_{1:H}; z_1, z_g).
$

Unrolling the optimized latent plan yields a sequence of intermediate latent subgoals at the high-level temporal resolution:
$
\tilde{z}_{i}
\;\triangleq\;
F^{(2)}(l^*_{1:i}; z_1),
\ i = 1, \ldots, H
$.

{Because macro-actions are latent rather than physical, their scale and support are determined by training rather than known a priori; we therefore use per-dimension statistics of macro-actions produced during training to initialize and bound the high-level search distribution (\cref{app:plan_params}).}

\textbf{Low-Level Planning.}
At execution time, the agent plans primitive actions to reach the first latent subgoal $\tilde{z}_1$. Starting from the current latent state $z_1$, we define a low-level energy function over a horizon of $h$ steps:
\[
\mathcal{E}_1(\hat{a}_{1:h}; z_1, \tilde{z}_1)
\;\triangleq\;
\left\| \tilde{z}_1 - F^{(1)}(\hat{a}_{1:h}; z_1) \right\|_1
\]
The optimal low-level action sequence is obtained via
$
a^*_{1:h}
=
\arg\min_{\hat{a}_{1:h}}
\mathcal{E}_1(\hat{a}_{1:h}; z_1, \tilde{z}_1)
$. {The agent executes the actions corresponding to the first $k$ low-level prediction steps before replanning. At successive replanning steps, we progressively shorten the high-level planning horizon (\cref{app:plan_params}).}

\section{Empirical Evaluation of Hierarchical Planning Across World Models}

\begin{figure*}[t]
\vspace{-1mm}
  \centering
  \includegraphics[width=\textwidth]{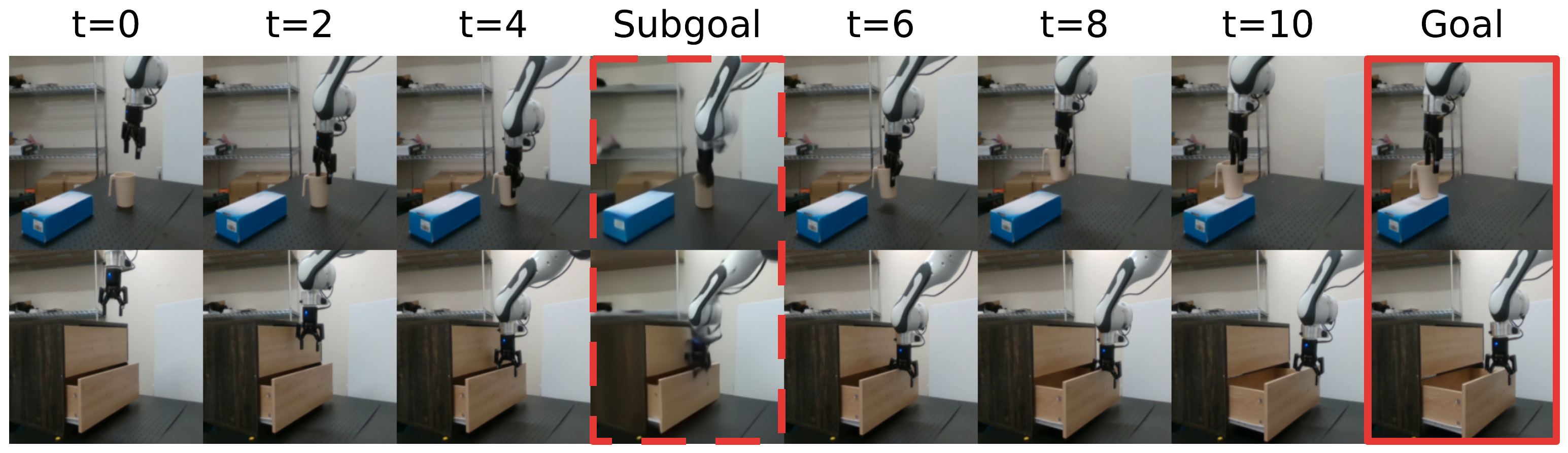}
  \caption{Executions on pick-\&-place and drawer manipulation with a dual-gripper Franka. Dotted red columns indicate subgoals inferred by the high-level planner and decoded for visualization. Executions on additional tasks are shown in \cref{app:franka_rollouts}.}
  \label{fig:robot_execution_main}
\vspace{-3mm}
\end{figure*}

We evaluate HWM on three latent world-model architectures — VJEPA2-AC \citep{assran2025v}, DINO-WM \citep{zhou2024dino}, and PLDM \citep{sobal2025learning} — under harder versions of their original settings: real-world Franka manipulation without explicit subgoals (VJEPA2-AC), Push-T extended from 25 to 75 timesteps (DINO-WM), and PointMaze with train–test layout mismatch and larger maps (PLDM).

\subsection{Franka Arm with Robotiq Gripper}
\label{subsec:robot}
We evaluate HWM on real-world multi-stage Franka manipulation: pick-\&-place and drawer opening/closing. These tasks require non-greedy behavior, where success depends on intermediate motions that temporarily increase distance to the goal.

\textbf{Platform and Task.}
We follow the VJEPA2-AC deployment setup and evaluate on real-world pick-\&-place and drawer manipulation using a 7-DoF Franka Emika Panda arm with a two-finger gripper. Pick-\&-place includes 10 start–goal pairs across two objects (cup and box); drawer manipulation covers 7 opening and closing tasks (\cref{app:robot_tasks}). We run N=5 independent trials per configuration, yielding 50 trials for pick-\&-place per object and 35 for drawer; per-configuration outcomes are highly consistent (\cref{app:per-config}).

\textbf{Baselines.}
\textbf{VJEPA2-AC} \citep{assran2025v} is a single-level, world-model-based planner. 
Because we use the same architecture and training data for our low-level world model, this comparison isolates the effect of hierarchical planning.
We additionally evaluate three vision–language-action models (VLA) baselines \textbf{Octo}, $\bm{\pi}_{0}$-FAST-DROID, and $\bm{\pi}_{0.5}$-DROID (see \cref{app:vla} for details).


\textbf{Data.}
World models are trained on approximately 130 hours of unlabeled real-robot manipulation data from
DROID \citep{khazatsky2024droid} and RoboSet \citep{kumar2023robohive}.
Observations include RGB images and end-effector proprioception; actions correspond to end-effector
delta poses (see \cref{app:franka} for details).
 
\textbf{Hierarchical Planning Setup.}
The agent receives current and goal observations, each specified by RGB images and end-effector poses, and outputs end-effector delta poses mapped to joint displacements via inverse kinematics. The high-level planner optimizes 4-D macro-actions (we ablate this choice in \cref{app:subgoals}), while the low-level planner optimizes primitive actions; both use CEM and replan every step. Full hyperparameters are in \cref{app:plan_params}..

\textbf{World Model Architectures.}
We follow the VJEPA2-AC training recipe for the low-level world model (see \cref{app:franka}). 
To train the high-level world model,
We use $N=3$ waypoint states sampled from trajectory segments spanning up to $4$ seconds, with the middle waypoint chosen uniformly at random. Macro-actions are encoded from the intervening low-level actions using a transformer-based action encoder, with the CLS token projected to the macro-action space.
For visualization, we decode latent representations into RGB images using a ViT decoder; these reconstructions are used solely for qualitative analysis and are not involved in planning or training (details in \cref{app:franka_decoder}).


\textbf{Main Results.} 
We report the findings in \cref{tab:franka}. Prior work with VJEPA2-AC \citep{assran2025v} has shown strong

\begin{wraptable}{r}{0.45\columnwidth}
\small
\centering
\renewcommand{\arraystretch}{1.2}
\setlength{\tabcolsep}{3pt}
\vspace{0pt}
\caption{\textbf{Zero-shot robotic manipulation with Franka}, evaluated end-to-end
from a single goal image. Each rate is computed over $N=5$ trials per
(start, goal) configuration; per-configuration breakdowns and 95\%
Clopper--Pearson confidence intervals are reported in  Table~\ref{tab:franka_per_config}. The
flat VJEPA2-AC planner {fails on tasks with non-greedy objectives} (pick-\&-place,
multi-stage drawer); HWM solves them without manual decomposition. }
\label{tab:franka}
\begin{tabular}{@{}l c c c@{}}
\toprule
Method & P\&P Cup & P\&P Box & Drawer \\
\midrule
Octo                 & 0\%     & 0\%     & 43\%  \\
$\pi_{0}$-FAST-DROID & 52\%   & 18\%    & --   \\
$\pi_{0.5}$-DROID    & 68\%   & 36\%    & --   \\
VJEPA2-AC            & 0\%   & 0\%    & 30\%  \\
\midrule
\shortstack[l]{VJEPA2-AC\\(hierarchy)} & \textbf{70\%  } & \textbf{60\% } & \textbf{70\% } \\
\bottomrule
\end{tabular}

\vspace{-9pt}
\end{wraptable}

 zero-shot generalization to novel environments for {simple tasks such as grasping or reaching, where greedy goal matching suffices}. Consistent with this, the flat planner succeeds on 2/7 drawer tasks admitting linear push/pull motions, but fails entirely on pick-\&-place and on {drawer variants requiring non-greedy, multi-stage planning}. In contrast, hierarchical planning solves both pick-\&-place and drawer tasks end-to-end from a single goal image. It generalizes zero-shot to novel environments and objects, achieving 70\% success on pick-\&-place cup and 60\% on box, and {solves drawer tasks requiring non-greedy, multi-stage planning} (e.g., moving down before translating laterally). Example executions are in \cref{fig:robot_execution_main} and \cref{app:franka_rollouts}. HWM failures arise mainly from perceptual imprecision (e.g., depth errors) or near-miss executions, suggesting that coarse high-level subgoals can lose details needed for precise low-level guidance.

We also compare against vision--language--action (VLA) baselines, noting that the comparison spans different goal-specification interfaces: HWM and Octo use a single goal \emph{image}, while the $\pi$-models use \emph{language} goals (\cref{app:vla}). HWM is competitive with these baselines despite significantly less training data ($\sim 77\times$), and far exceeds Octo under identical image-goal supervision.

\begin{conclusionbox}
\textbf{Takeaway:} HWM enables zero-shot {manipulation under non-greedy objectives} on a real robot from a single goal image, where flat world-model planners fail.
\end{conclusionbox}

\subsection{Push-T}
\label{sec:push-t}

We instantiate HWM on top of DINO-WM \citep{zhou2024dino} and evaluate it on goal-conditioned Push-T, where an agent must push a T-shaped object to a goal configuration sampled from a validation trajectory. To stress-test long-horizon planning, we evaluate start–goal pairs up to $d=75$ timesteps apart, versus $d=25$ evaluated in DINO-WM. Models are trained from the dataset released with DINO-WM.

\textbf{Baselines.}
We compare against (i) DINO-WM~\citep{zhou2024dino}, the flat-planning baseline; (ii) HIQL~\citep{park2023hiql}, a state-of-the-art \emph{hierarchical} goal-conditioned RL method; (iii) GCIQL~\citep{kostrikov2021offline} and HILP~\citep{park2024foundation} as additional goal-conditioned and zero-shot RL baselines (\cref{app:gcrl}).

\textbf{Hierarchical Planning Setup.}
Hierarchical planning is performed using CEM at two temporal scales, with a high-level planner optimizing macro-actions and a low-level planner refining primitive actions in a receding-horizon manner. Full CEM hyperparameters are provided in \cref{app:plan_params}.

\begin{wraptable}{r}{0.48\columnwidth}
\vspace{-10pt}
\small
\centering
\renewcommand{\arraystretch}{1.2}
\setlength{\tabcolsep}{4pt}

\caption{\textbf{Push-T Performance Across Task Horizons}. In each trial, a random start–goal pair $d$ steps apart is sampled from a validation trajectory.}
\label{tab:push_t}
\begin{tabular*}{\linewidth}{@{\extracolsep{\fill}} l @{\hspace{3pt}} c c c @{}}
\toprule
Method & $d=25$ & $d=50$ & $d=75$ \\
\midrule
GCIQL    & $40\%$ & $25\%$ & $7.5\%$ \\
HIQL     & $55\%$ & $30\%$ & $20\%$ \\
HILP     & $25\%$ & $13\%$ & $0\%$ \\
DINO-WM  & $84\%$ & $55\%$ & $17\%$ \\
\midrule
DINO-WM (hier.) & \textbf{89\%} & \textbf{78\%} & \textbf{61\%} \\
\bottomrule
\end{tabular*}

\vspace{-10pt}
\end{wraptable}

\begin{figure}[t]
\vspace{-2mm}
\centering

\begin{subfigure}{0.48\linewidth}
\centering
\includegraphics[width=\linewidth]{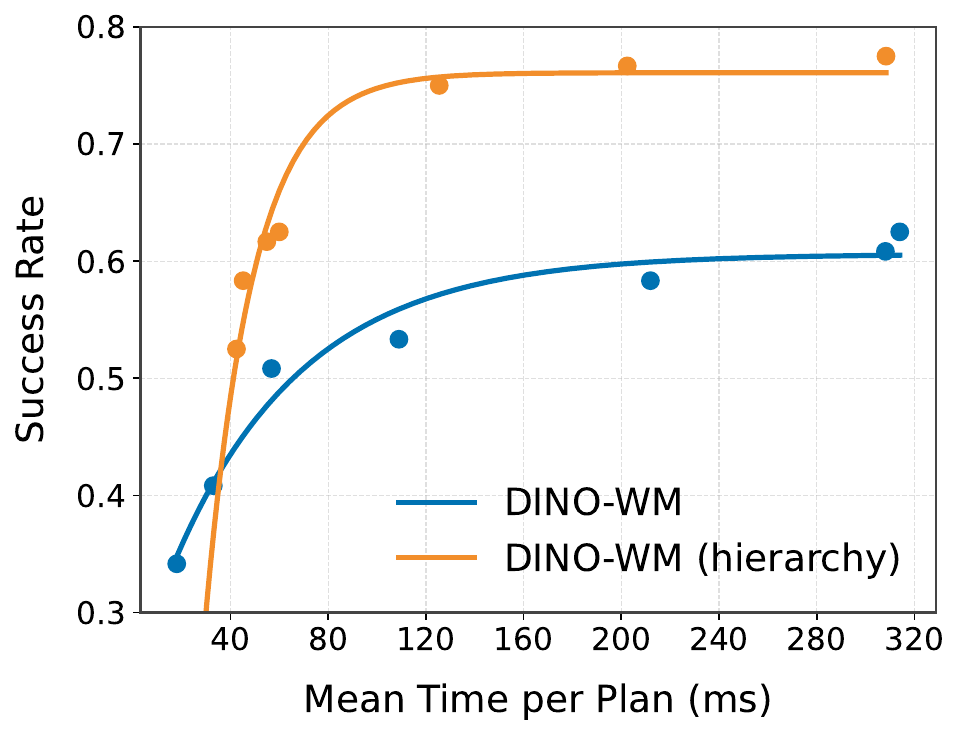}
\end{subfigure}
\hfill
\begin{subfigure}{0.48\linewidth}
\centering
\includegraphics[width=\linewidth]{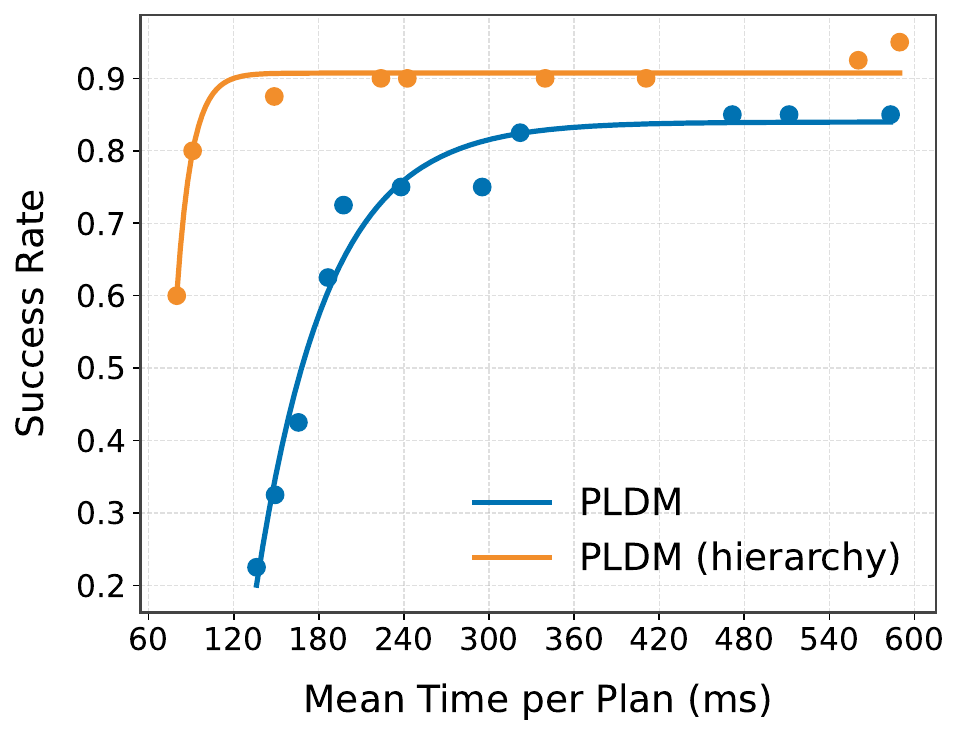}
\end{subfigure}

\caption{\textbf{Success rate vs.\ test-time compute.} We vary the planning budget by sweeping CEM/MPPI samples and iterations (see \cref{app:comp_study}). 
Hierarchical planning achieves higher success while using up to $3\times$ less test-time compute than flat planners.
\textbf{Left:} Push-T success rate vs.\ planning time ($d=50$).
\textbf{Right:} Diverse maze success rate vs.\ planning time ($D\!\in\![9,12]$).
}

\label{fig:compute_analysis}
\end{figure}

\textbf{Main Results.}
As shown in \cref{tab:push_t}, hierarchical planning consistently outperforms DINO-WM as task difficulty increases. While the flat planner’s success rate drops sharply as the task horizon grows, the hierarchical planner remains substantially more robust at longer horizons. The performance gap is not explained by parameter count: capacity-matched flat planners do not match HWM and sometimes degrade with increased capacity (\cref{app:capacity}). Hierarchical planning also exhibits a more favorable compute–performance trade-off, capable of achieving higher success rates while requiring $3 \times$ less compute per planning step compared to the flat planner (\cref{fig:compute_analysis}). These results highlight the advantage of planning over temporally abstract macro-actions, which reduces optimization complexity while preserving effective long-horizon control. Moreover, policy baselines (GCIQL, HIQL, HILP) performances drop sharply at longer horizons, suggesting limited robustness to long-horizon generalization without explicit planning. See \cref{fig:pusht_execution} for examples of executions and subgoals.

\begin{conclusionbox}
\textbf{Takeaway}: HWM improves long-horizon control while reducing inference-time compute.
\end{conclusionbox}

\subsection{Diverse Maze}
We instantiate HWM on top of PLDM \citep{sobal2025learning} and evaluate it on Diverse Maze navigation, focusing on zero-shot generalization to larger, unseen layouts that require long-horizon planning.

\textbf{Platform and Task.} 
We build on the MuJoCo PointMaze environment~\citep{conf/iros/TodorovET12}, using top-down RGB renderings as input. Maze layouts are randomly generated on a $10\times10$ grid with connected free space; models are trained on 25 layouts and evaluated on 20 held-out layouts. Start and goal locations are sampled uniformly with grid-distance separation $D$, defining easy ($D\!\in\![5,8]$), medium ($D\!\in\![9,12]$), and hard ($D\!\in\![13,16]$) regimes. Dataset and training details are provided in \cref{app:diverse_maze}.

\textbf{Planning and Baselines.}
Hierarchical planning uses MPPI at two temporal scales (details in \cref{app:diverse_maze}). We compare against the flat PLDM planner, goal-conditioned RL baselines GCIQL and HIQL \citep{park2023hiql}, and the zero-shot RL method HILP \citep{park2024foundation}.

\begin{wraptable}[13]{r}{0.48\columnwidth}
\small
\vspace{-10pt}
\centering
\renewcommand{\arraystretch}{1.2}
\setlength{\tabcolsep}{4pt}

\caption{\textbf{Diverse Maze Performance Across Task Horizons}. Agents are evaluated zero-shot on navigation tasks in held-out test maps. For each task, start and goal positions are randomly sampled such that their shortest-path distance is $D$.}
\label{tab:maze}
\begin{tabular*}{\linewidth}{@{\extracolsep{\fill}} l @{\hspace{3pt}} c c c @{}}
\toprule
Method & $D\!\in\![5,8]$ & $D\!\in\![9,12]$ & $D\!\in\![13,16]$ \\
\midrule
GCIQL    & 85\% & 40\% & 33\% \\
HIQL     & 88\% & 73\% & 48\% \\
HILP     & 48\% & 20\% & 10\% \\
PLDM     & 100\% & $84\%$ & $44\%$ \\
\midrule
PLDM (hier.) & \textbf{100\%} & \textbf{95\%} & \textbf{83\%} \\
\bottomrule
\end{tabular*}

\vspace{-10pt}
\end{wraptable}

\textbf{Main Results.}
As shown in \cref{tab:maze}, the performance gap between HWM and PLDM widens as task horizon increases. Even in settings where the flat planner performs reasonably well, it requires substantially higher test-time compute to match HWM performance (\cref{fig:compute_analysis}). In contrast, HWM achieves higher success with $4\times$ less compute. Both single-level and hierarchical PLDM planners outperform goal-conditioned and zero-shot RL baselines on held-out layouts, highlighting the robustness of model-based planning to geometric distribution shift. See \cref{fig:maze_plans} for high-level plan visualizations.

\begin{conclusionbox}
\textbf{Takeaway:} HWM consistently improves zero-shot planning across diverse latent world models.
\end{conclusionbox}

\Needspace{12\baselineskip}
\subsection{Ablations and Analysis}
\label{sec:ablations}

{%
\setlength{\intextsep}{0pt}
\begin{wrapfigure}{r}{0.48\columnwidth}
\centering
\includegraphics[width=\linewidth,trim=0 0 0 7.2bp,clip]{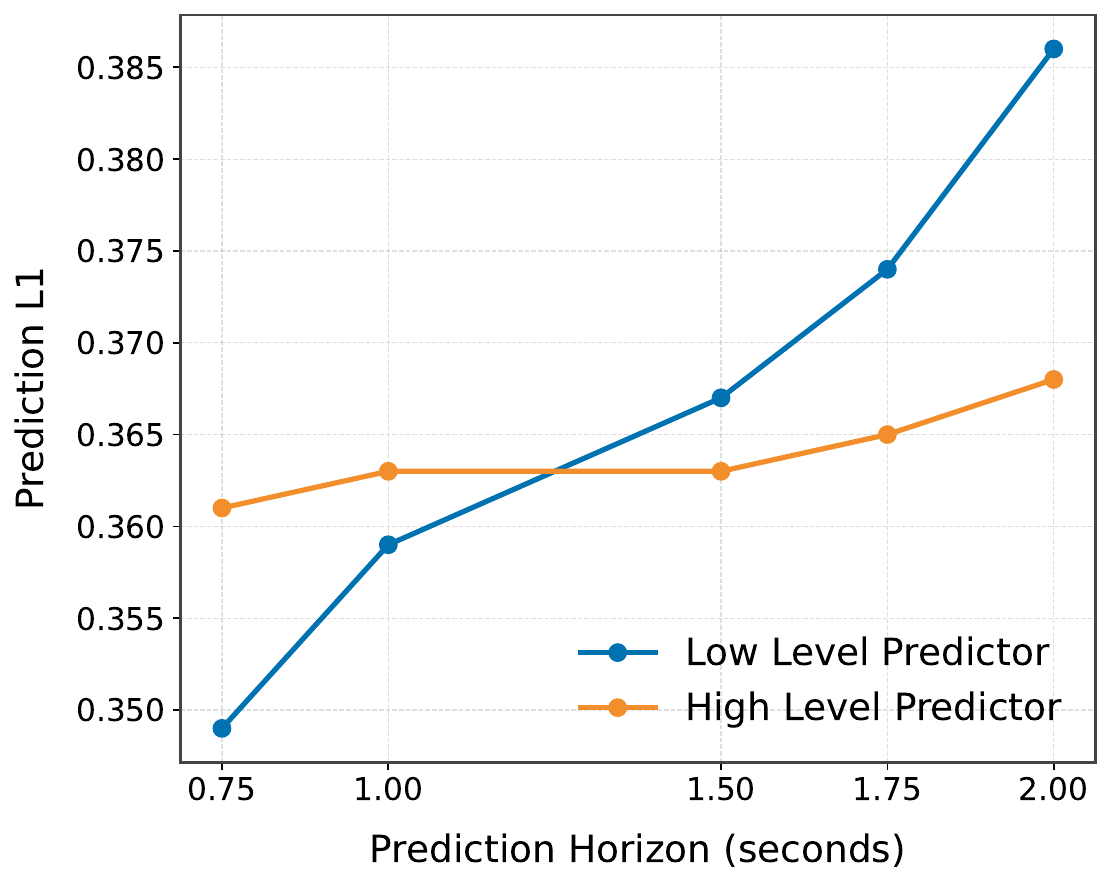}
\vspace{-20pt}

\caption{\textbf{Prediction error ($L_1$) as a function of prediction horizon.}
For short horizons ($\leq$ 1\,s), the low-level world model is more accurate.
For longer horizons ($\geq$ 1.5\,s), a single-step prediction from the high-level world model outperforms autoregressive rollouts of the low-level model, reflecting reduced error accumulation.}

\label{fig:latent_prediction_l1_vs_l2}
\end{wrapfigure}

\textbf{Do high-level world models improve long-horizon prediction?} A key challenge in long-horizon planning with learned world models is error accumulation, where small one-step errors compound over autoregressive rollouts. We hypothesize that longer-temporal-scale models improve long-horizon predictions by requiring fewer autoregressive steps. To test this, we condition low- and high-level world models trained on DROID on held-out initial observations and predict future states up to 2 seconds ahead, measuring $\ell_1$ error to the ground-truth future latent state. Predicting 2 seconds ahead requires up to 16 low-level autoregressive steps, whereas the high-level model produces a single-step prediction. As shown in \cref{fig:latent_prediction_l1_vs_l2}, the low-level model is more accurate for short horizons ($\leq$ 1,s), while the high-level model achieves lower error for longer horizons ($\geq$ 1.5,s), supporting a hierarchy where high-level planning provides long-term guidance and low-level planning handles short-term precision.\par
}

\begin{table}[htbp]
\centering
\small
\setlength{\tabcolsep}{4pt}
\caption{{\textbf{Learned macro-actions versus alternative high-level action representations.} \textbf{(a)} Concatenated primitive actions: planning success rates on medium/hard Diverse Maze tasks. \textbf{(b)} Handcrafted delta-pose actions, which summarize each action sequence by its net end-effector displacement: alignment of the offline plan with expert behavior on Franka (mean $\pm$ SE), measured by cosine similarity and $\ell_1$ distance.}}
\label{tab:macro-action-ablation}
\begin{subtable}[t]{0.47\linewidth}
\centering
\caption{{Concatenated actions: Diverse Maze.}}
\label{tab:latent-vs-concat}
{%
\begin{tabular}{@{}lcc@{}}
\toprule
Action & $D\!\in\![9,12]$ & $D\!\in\![13,16]$ \\
\midrule
Concatenated & 52\% & 37\% \\
Macro-actions & \textbf{95\%} & \textbf{83\%} \\
\bottomrule
\end{tabular}}
\end{subtable}\hfill
\begin{subtable}[t]{0.50\linewidth}
\centering
\caption{{Handcrafted actions: Franka.}}
\label{tab:action_encoder}
{%
\begin{tabular}{@{}lcc@{}}
\toprule
Action & Cos $\uparrow$ & $\ell_1$ $\downarrow$ \\
\midrule
Delta pose & $0.80\pm0.02$ & $0.088\pm0.005$ \\
Macro-actions & $\mathbf{0.88\pm0.03}$ & $\mathbf{0.080\pm0.002}$ \\
\bottomrule
\end{tabular}}
\end{subtable}

\end{table}

{\textbf{Role of learned macro-actions.} A key question is how to parameterize macro-actions in the hierarchical world model. We compare learned latent macro-actions—obtained by encoding sequences of low-level actions—against concatenated primitive actions on Diverse Maze and hand-crafted net end-effector displacement (delta-pose) summaries on Franka. Learned macro-actions consistently outperform both. Compared to concatenating primitive actions, macro-actions achieve substantially higher planning performance (\cref{tab:latent-vs-concat}) by compressing action sequences into a lower-dimensional space, reducing the high-level planning search complexity. On Franka, macro-actions also outperform delta-pose actions, producing offline plans that better align with expert trajectories (\cref{tab:action_encoder}). This is likely because delta-pose summaries collapse temporally extended, non-greedy trajectories into a single displacement, while macro-actions retain information needed to predict feasible transitions.\par}

\section{Related Works}

\textbf{World models and planning.} Learning predictive models of dynamics is a core idea in model-based RL~\citep{sutton1981adaptive,deisenroth2011pilco,ha2018worldmodels,hafner2023mastering,hansen2023td}, with recent work training latent dynamics from pixels for planning and policy optimization~\citep{ebert2017self,hafner2019learning,schwarzer2020data,2026tdjepa}. Closely related to our setting, latent world models trained on reward-free offline data via next-latent prediction generalize zero-shot to downstream control by simulating action outcomes optimizing costs~\citep{assran2025v,sobal2025learning,terver2026drivessuccessphysicalplanning,zhou2024dino,goswami2025world,maes2026leworldmodel}. Long-horizon planning with such models is bottlenecked by compounding prediction errors~\citep{talvitie2014model,janner2019trust} and exponential search complexity~\citep{ichter2020broadly}. A contemporary line of work instead couples video generation with action prediction in a single model — sometimes called World Action Models — leveraging pretrained video diffusion backbones to inherit visuomotor priors \citep{hu2024video,liao2025genie,li2025unified,ye2026world}; these methods learn an explicit policy, whereas latent world model planners like HWM learn only dynamics and perform zero-shot MPC at inference.

\textbf{Hierarchical Control.} Hierarchical RL approaches use options, skills, or subgoal-conditioned policies for temporal abstraction~\citep{sutton1999between,bacon2017option,hafner2022deep,park2023hiql,gumbsch2023learning,schiewer2024exploring,hansen2024hierarchical}, but require reward supervision and are limited to the training task distribution. Hierarchical MPC in optimal control~\citep{fang2019dynamics,li2021planning,kogel2025safe} is task-agnostic but largely restricted to low-dimensional state or known dynamics. Table~\ref{tab:hierarchical_comparison} summarizes the design axes of previous hierarchical model-based methods and how HWM differs from them. A complementary line of work focuses on unsupervised skill or subgoal discovery \citep{zhang2021hierarchical,mazzaglia2022choreographer,zhang2025efficient}. We view this as orthogonal: HWM uses random or fixed-stride waypoint sampling, and structured skill discovery could in principle replace this without changing the rest of the framework.

\begin{table}[h]
\centering
\footnotesize
\setlength{\tabcolsep}{5pt}
\caption{\textbf{Comparison of hierarchical model-based methods.} HWM is the only entry supporting zero-shot MPC directly from pixels, using world models pretrained via next-latent prediction, with predictions from coarser models serving as subgoals for finer-scale MPC via latent matching.}

\label{tab:hierarchical_comparison}
\begin{tabular}{@{}l@{\hspace{2pt}}c l l l@{}}
\toprule
\textbf{Method} & \textbf{Zero-shot} & \textbf{Representation} & \textbf{Training} & \textbf{Hierarchical Interface} \\
\midrule
\textbf{HWM (ours)} & \checkmark & pixels $\rightarrow$ latent & Next-latent pred. & \textbf{Subgoal matching (latent)} \\
\midrule
CAVIN \citep{fang2019dynamics} & \checkmark & structured state & Learned dynamics & Subgoal matching (state) \\
LAT \citep{li2021planning} & \checkmark & structured state & Learned dynamics & Skill-conditioned policy \\
Safe HMPC \citep{kogel2025safe} & $\times$ & structured state & Analytical dynamics & Reference tracking \\
\midrule
IQL-TD-MPC \citep{chitnis2024iql} & $\times$ & state $\rightarrow$ latent & Offline RL + MPC & Goal-conditioned policy \\
Director,Puppeteer \citep{hafner2022deep,hansen2024hierarchical} & $\times$ & pixels $\rightarrow$ latent & Model-based RL & Goal-conditioned policy \\
THICK \citep{gumbsch2023learning} & $\times$ & pixels $\rightarrow$ latent & MB-RL + reward & Subgoal matching (latent) \\
\bottomrule
\end{tabular}

\vspace{-1em}
\end{table}

\vspace{-2mm}

\section{Conclusion}
We presented HWM, a hierarchical MPC framework for zero-shot planning with latent world models trained from offline trajectories. By learning dynamics at multiple temporal scales in a shared latent space, HWM lets high-level predictions serve as subgoals for short-horizon low-level planning, without hierarchical policies, task rewards, or manual decomposition. On a real Franka robot, this enables pick-and-place and multi-stage drawer manipulation from a single goal image, where the corresponding flat planner fails. Across all task settings, the same principle improves long-horizon success and reduces planning cost across multiple world-model backbones. These results suggest that hierarchy is a practical mechanism for extending latent world-model MPC beyond short, greedy behaviors.

\section{Limitations}


Despite these gains, success rates degrade with task horizon for every method we evaluate; closing this gap will require further progress in world modeling and planning. First, HWM uses a single latent space across hierarchy levels: while this enables direct subgoal transfer, longer horizons and more open-ended environments may benefit from coarser, level-specific abstractions. Second, subgoal quality is shaped by choices at training and inference. At training time, we supervise the high-level world model with randomly sampled or fixed-stride waypoints; skill or subgoal discovery~\citep{zhang2021hierarchical,mazzaglia2022choreographer,zhang2025efficient}, or dynamic chunking~\citep{hwang2025dynamic}, could replace this waypoint scheme without changing the rest of the framework, potentially yielding better subgoals.
At inference time, incorporating feedback from the lower-level planner into high-level subgoal optimization may help compensate for the loss of detail in high-level predictions needed to guide precise low-level control.

\clearpage
\bibliography{references}

\newpage

\appendix
\crefalias{section}{appendix}
\crefalias{subsection}{appendix}
\onecolumn
\startcontents[appendix]

\section*{Appendix Contents}
\setcounter{tocdepth}{2} 
\printcontents[appendix]{}{1}{}

\newpage

\section{Training Details}
\label{app:training}

\subsection{Low Level World Model} \label{app:l1_wm}

We train the low-level latent world model $F^{(1)}_{\theta}$ to approximate the true transition dynamics $s_{t+1} = f(s_t, a_t)$. Given a trajectory sequence $\tau = (s_1,a_1,s_2,a_2,\dots,a_{T-1},s_T)$, we first encode the states independently to obtain a sequence of latent features $(z_k)_{k\in [T]}$.  We feed the interleaved sequence of actions and latent features $(a_k, z_k)_{k \in [T-1]}$ into the world model to obtain a sequence of next state representation predictions $(\hat{z}_{k+1})_{k \in [T-1]}$. The teacher-forcing loss is computed as:
\begin{equation}
\begin{aligned}
\hat{z}_{k+1} &:= F^{(1)}_{\theta}\!\left((a_t, z_t)_{t \le k}\right) \\
\mathcal{L}_{\mathrm{tf}}(\theta)
&:= \frac{1}{T-1}\sum_{k=1}^{T-1}\left\|\hat{z}_{k+1}-z_{k+1}\right\|_1 .
\end{aligned}
\label{eq:l1_losses}
\end{equation}
\vspace{-1em}

To mitigate compounding prediction errors over multiple steps, we additionally compute a multi-step autoregressive rollout loss. Let $F^{(1)}_\theta(a_{k:j}; z_k)$ denote the predicted representation of $s_{j+1}$ obtained by autoregressively running the world model with actions $a_{k:j}$, starting from $z_k$, where $1 \leq k \leq j \leq T-1$. The multi-step rollout loss over up to $T-1$ prediction steps is:
\vspace{-1mm}
\begin{align}
    \mathcal{L}_{\mathrm{roll}}(\theta) := 
    \sum_{j=2}^{T-1}
    \|F^{(1)}_{\theta}(a_{1:j}; z_1) - z_{j+1} \|_1
\end{align}

The overall training objective is given by a weighted sum of the two quantities:
\begin{align}
    L(\theta) := \gamma_{tf} * \mathcal{L}_{\mathrm{tf}}(\theta) + \gamma_{roll} *\mathcal{L}_{\mathrm{roll}}(\theta)
\end{align}

\subsection{Franka Arm with Robotiq Gripper}\label{app:franka}

\textbf{Data.} We use approximately 96 hours of unlabeled videos from the DROID dataset \citep{khazatsky2024droid} and 30 hours from RoboSet \citep{kumar2023robohive}.
Both datasets consist of short video clips of manipulation trajectories collected from real robots.
Observations include RGB images and end-effector proprioception; actions correspond to end-effector
delta poses.

\textbf{State, Actions, and Preprocessing.}
Videos are sampled at a spatial resolution of $256 \times 256$ and with a frame rate uniformly sampled from $(3, 10)$ FPS for DROID and $(1, 5)$ FPS for RoboSet. At timestep $k$, the state is defined as $s_k = (x_k, p_k)$, where $x_k$ is the raw RGB observation and
$p_k \in \mathbb{R}^7$ is the end-effector state expressed in the robot base frame.
The first three dimensions of $p_k$ correspond to Cartesian position, the next three correspond to orientation represented as extrinsic Euler angles, and the final dimension represents the gripper state. Actions are defined as differences in end-effector state between consecutive frames: $a_k = p_{k+1} - p_k$. We apply random resize-and-crop augmentations to video clips, with the aspect ratio sampled uniformly from $(0.75, 1.33)$.

\textbf{Low Level World Model.}
Following the setup of VJEPA2-AC \citep{assran2025v}, we sample fixed-length trajectories containing $T=16$ observation frames $\{(p_k, x_k)\}_{k=1}^{16}$ and 15 intervening actions $\{a_k\}_{k=1}^{15}$.
Each image $x_k$ is encoded using the pretrained, frozen ViT-g/16 backbone, producing latent feature maps $(z_k)_{k=1}^{16}$.
The world model is instantiated as a $\sim$300M-parameter ViT.
It takes temporally interleaved sequences of latent features, end-effector states, and actions
$\{(z_k, p_k, a_k)\}_{k=1}^{15}$ and produces 15 next-step latent predictions $\{\hat{z}_{k}\}_{k=2}^{16}$, with targets $z_{2:16}$. We refer to the VJEPA2 paper for full architectural details.

\textbf{High-Level World Model.}
To train the high-level world model, we sample variable-length trajectory segments $\{(a_k, p_k, x_k)\}_{k \in T}$ spanning $(0.33, 4)$ seconds and select $N=3$ waypoint indices per segment, with the middle waypoint sampled uniformly. The high-level model shares the ViT architecture of the low-level model, but is conditioned on latent macro-actions rather than primitive actions. Conditioned on the current and previous waypoints and macro-actions, it is trained via teacher forcing to predict the next waypoint. Latent macro-actions are produced by a transformer-based action encoder $A_\psi$, whose CLS token is mapped to a macro-action by an MLP head.

\begin{table}[h]
\centering
\small
\renewcommand{\arraystretch}{1.1}
\setlength{\tabcolsep}{4pt}
\caption{\textbf{Training hyperparameters for Droid + Roboset.} $\gamma_{tf}$ and $\gamma_{roll}$ denote the coefficients for teacher-forcing and rollout losses, respectively, and context $T$ counts observation frames, including the initial frame. All other hyperparameters follow those used by VJEPA2-AC.}
\label{tab:franka_train_hyperparams}
\begin{tabular}{lcccccc}
\toprule
World Model & Epochs & Batch Size & $\gamma_{tf}$ & $\gamma_{roll}$ & context $T$ \\
\midrule
VJEPA2-AC (Low-Level) & 200 & 256 & 1.0 & 1.0 & 16 \\
VJEPA2-AC (High-Level) & 120 & 768 & 1.0 & 0.0 & 3 \\
\bottomrule
\end{tabular}%

\vspace{-1em}
\end{table}

\subsection{Push-T}\label{app:pusht_train}

\textbf{Data.}
We use the offline dataset from DINO-WM, which consists of 18500 trajectories. Each trajectory consists of a sequence of $(x_t, a_t, p_t)$, where $x_t$ is an RGB observation, $a_t$ is a 2D action vector, and $p_t$ is the proprioceptive state denoting current agent position and velocity.

\textbf{Low Level World Model.}
We use DINO-WM \citep{zhou2024dino} as our low-level world model. The encoder is a pretrained DINOv2 network \citep{oquab2023dinov2}, which is kept frozen during training. The world model is a 25M params ViT with causal masking. Training samples contain $T=4$ frames at stride 5, spanning 15 raw timesteps and providing three next-state targets $z_{2:4}$. The action $a_k$ is formed by concatenating the 5 primitive actions executed between $k$ and $k+1$, yielding a $5 \times 2 = 10$–dimensional action vector.
Each latent patch $z_k^i$ attends only to corresponding earlier patches $z_j^i$, $1 \leq j < k$, within the sample. Actions are first embedded using an MLP and then concatenated to each patch vector $z_k^i$. Proprioceptive inputs are incorporated using the same embedding and concatenation scheme. We refer to the original DINO-WM paper for full architectural details.

\textbf{High-Level World Model.}
To construct training sequences, we subsample trajectory segments with lengths uniformly drawn between 25 and 70 timesteps. From each segment, we sample $N=5$ waypoint states, which define the high-level transitions.
The high-level world model is a scaled-up variant of the low-level world model (25M $\rightarrow$ 75M parameters), obtained by increasing the number of layers from 6 to 10, the embedding dimension from 384 to 768, and the MLP dimension from 2048 to 3072, while reducing the number of attention heads from 16 to 12. This increased capacity enables more effective modeling of long-horizon dynamics.
Latent macro-actions are produced by a transformer-based action encoder $A_\psi$, where the CLS token is passed through an MLP head to obtain the macro-action representation. See \cref{tab:pusht_train_hyperparams} for hyperparameters.

\begin{table}[h]
\centering
\small
\renewcommand{\arraystretch}{1.1}
\setlength{\tabcolsep}{4pt}
\caption{\textbf{Training hyperparameters for Push-T.} $\gamma_{tf}$ and $\gamma_{roll}$ denote the coefficients for teacher-forcing and rollout losses, respectively, and context $T$ counts frames including the initial observation. All other hyperparameters follow those used by DINO-WM on Push-T.}
\label{tab:pusht_train_hyperparams}
\begin{tabular}{lcccccc}
\toprule
World Model & Epochs & Batch Size & $\gamma_{tf}$ & $\gamma_{roll}$ & context $T$ \\
\midrule
DINO-WM (Low-Level) & 100 & 256 & 1.0 & 0.0 & 4 \\
DINO-WM (High-Level) & 500 & 128 & 1.0 & 0.0 & 5 \\
\bottomrule
\end{tabular}%

\vspace{-1em}
\end{table}

\subsection{Diverse Maze}\label{app:diverse_maze}

\textbf{Environment.} The environment is a variant of Mujoco PointMaze \citep{conf/iros/TodorovET12}, which contains a point mass agent with a 4D state vector $(\mathrm{global} \ x, \mathrm{global} \ y, v_x, v_y)$, where $v$ is the agent velocity. The model receives a top down view of the maze rendered as $(98,98,3)$ RGB image tensor as input, instead of relying on $(\mathrm{global} \ x, \mathrm{global})$ directly. Mujoco PointMaze allows customizing the maze layout via a grid, where each cell is a wall or space. We opt for a $8 \times 8$ grid (excluding outer wall). Maze layouts are generated randomly. Only the following constraints are enforced: 1) all the space cells are interconnected, 2) percentage of space cells range from $50\%$ to $80\%$.

\textbf{Data.}
Each episode is initialized by placing the agent at a random $(x, y)$ location in the maze and sampling an initial velocity $(v_x, v_y)$. A trajectory is then generated by randomly sampling and executing actions for 100 steps, using an action repeat of 4. We collect 2,000 episodes per map and use 25 maps for training; resulting in a total of 5M transitions.

\textbf{Low Level World Model.}
We train the low-level world model following the PLDM recipe \citep{sobal2025learning}, jointly learning the encoder and world model from scratch using offline trajectories. 
Each training sample consists of 15 observation frames and the 14 intervening actions subsampled from a trajectory.
The observation sequence is encoded by a convolutional network into spatial latent features $z_{1:15}$. 
Conditioned on the initial latent $z_1$ and the action sequence $a_{1:14}$, a convolutional world model autoregressively produces 14 future predictions $\hat{z}_{2:15}$, excluding the initial state.
To prevent representational collapse, we apply VICReg~\citep{bardes2021vicreg} to the latent features $z_{1:15}$, following the original PLDM formulation. Refer to \cref{tab:maze_train_hyperparams} for hyperparameters.
We use the same PLDM architecture for Diverse Maze, comprising a lightweight convolutional encoder (33k params) and predictor (20k params); we refer to the original paper for full details.

\textbf{High-Level World Model.}
To construct training sequences, we sample trajectory segments spanning 60 timesteps and extract 7 waypoint states, including both endpoints, using a fixed stride of 10 timesteps. These define 6 high-level transitions. Primitive action chunks are encoded into an 8-dimensional macro-action via an MLP. We reuse the backbone of a pretrained low-level world model and keep it frozen, training only the high-level world model. The high-level model is a higher-capacity convolutional network, enabling long-horizon prediction. Conditioned on the initial waypoint latent \(z_1\) and the macro-action sequence \(l_{1:6}\), it autoregressively predicts the 6 subsequent waypoint latents. Hyperparameters are provided in \cref{tab:maze_train_hyperparams}.

\begin{table}[h]
\centering
\footnotesize
\renewcommand{\arraystretch}{1.1}
\setlength{\tabcolsep}{4pt} 
\caption{\textbf{Training hyperparameters for Diverse Maze.} $\alpha$, $\beta$, $\lambda$, and $\omega$ are parameters in the PLDM objective used to prevent representation collapse. $\mathcal{L}_{\text{proprio}}$ coef. denotes the coefficient on the proprioceptive prediction loss. $\gamma_{tf}$ and $\gamma_{roll}$ are the coefficients for teacher-forcing and rollout losses, and pred $T$ is the number of autoregressive prediction steps during training, excluding the initial state. For the high-level world model, we reuse the encoder trained at the low level and keep it frozen, eliminating additional representation learning.}
\label{tab:maze_train_hyperparams}
\begin{tabular}{lccccccccccc}
\toprule
World Model & Epochs & Batch Size & lr & $\alpha$ & $\beta$ & $\lambda$ & $\omega$ & $\gamma_{tf}$ & $\gamma_{roll}$ & pred $T$ & $\mathcal{L}_{\text{proprio}}$ \\
\midrule
PLDM (Low-Level) & 3 & 128 & 0.018 & 29.4 & 17.9 & 2.80 & 4.81 & 0.0 & 1.0 & 14 & 2.42 \\
PLDM (High-Level) & 5 & 128 & 0.018 & -- & -- & -- & -- & 0.0 & 1.0 & 6 & 1 \\
\bottomrule
\end{tabular}%

\vspace{-1em}
\end{table}

\section{Planning Details}
\label{app:plan_params}

For Franka and Push-T, we use the Cross-Entropy Method (CEM) \citep{rubinstein2004cross} for trajectory optimization in continuous action spaces. At each planning step, CEM maintains a factorized Gaussian over action sequences, iteratively sampling trajectories, evaluating them with the world model, and refitting to elite samples by updating the mean and variance. To improve stability and avoid premature collapse, variance updates are smoothed with an exponential moving average. After a fixed number of iterations, the optimized mean action sequence is executed, and planning proceeds in a receding-horizon manner. See \cref{tab:cem_l1} and \cref{tab:cem_l2} for the single-level and hierarchical planner parameters, respectively.

{\textbf{Macro-action search distribution.}
The low-level planner optimizes primitive actions, whose scale is fixed by the environment, and is initialized with zero mean and unit variance. Macro-actions, in contrast, live in a learned latent space whose scale and support are set by the action encoder $A_{\psi}$. We therefore record the macro-actions produced by $A_{\psi}$ during training in a fixed-size FIFO buffer holding the most recent $N=4096$ of them, stored alongside the model weights so that planning requires no access to the training data. At plan time we compute the per-dimension mean and standard deviation over this buffer and use them to initialize the high-level Gaussian, and we clip sampled macro-actions to the per-dimension $2\%$ and $98\%$ quantiles at every CEM iteration. This keeps high-level search within the latent-action region where $F^{(2)}_{\phi}$ was trained, without additional tuning.}

{\textbf{Planning horizon schedules.}
We measure elapsed time $t$, execution budget $B$, and replanning interval $k$ in low-level prediction steps. A replanning interval of $k$ means executing actions for the first $k$ low-level predictions before replanning. Each low-level prediction step spans five environment steps on Push-T and one environment step on Diverse Maze. At each replanning step, the standalone single-level planner caps its prediction horizon by the remaining budget:
\[
h_t=\min\!\left(h, B-t\right),
\]
where $h$ is the maximum prediction horizon reported in \cref{tab:cem_l1,tab:mppi_l1} and $h_t$ is the horizon used at replanning time $t$. In hierarchical planning, the low-level planner uses a fixed horizon $h$ to reach the next subgoal. The high-level horizon starts at $H$ and decreases by one whenever $h$ low-level steps have elapsed, subject to a minimum:
\[
H_t=\max\!\left(H_{\min}, H-\left\lfloor t/h\right\rfloor\right),
\qquad h_t=h.
\]
Here, $h$ and $H$ are the prediction horizons reported in \cref{tab:cem_l2,tab:mppi_l2}, while $h_t$ and $H_t$ denote the horizons used at replanning time $t$. We use $H_{\min}=1$ for Push-T and $H_{\min}=3$ for Diverse Maze.}

\begin{table}[htbp]
\centering
\small
\setlength{\tabcolsep}{5pt}
\renewcommand{\arraystretch}{1.15}

\captionsetup{position=top,skip=5pt}
\captionof{table}{{Single-level CEM planning hyperparameters for Franka and Push-T.}}
\label{tab:cem_l1}
\begin{tabular}{lccccccc}
\toprule
 & $k$ & {$B$} & \# elites & \# iters & \# samples & Var EMA & $h$ \\
\midrule
Franka & 1 & {--} & 20 & 15 & 2400 & 0.75 & 6 \\
Push-T ($d=25$) & 5 & {10} & 30 & 30 & 300 & 0 & 5 \\ 
Push-T ($d=50$) & 5 & {20} & 10 & 30 & 1200 & 0.9 & 10 \\
Push-T ($d=75$) & 5 & {25} & 20 & 20 & 1000 & 0 & 15 \\
\bottomrule
\end{tabular}
\par\vspace{10pt}

\setlength{\tabcolsep}{2pt}
\captionsetup{position=top,skip=5pt}
\captionof{table}{{Hierarchical CEM planning hyperparameters for Franka and Push-T.}}
\label{tab:cem_l2}
\begin{tabular}{l@{\hspace{2pt}}cccccccccccc}
\toprule
 & & & \multicolumn{5}{c}{High-level planner} & \multicolumn{5}{c}{Low-level planner} \\
\cmidrule(lr){4-8}\cmidrule(lr){9-13}
 & $k$ & {$B$} & \# elites & \# iters & \# samples & Var EMA & $H$
 & \# elites & \# iters & \# samples & Var EMA & $h$ \\
\midrule
Franka & 1 & {--} & 22 & 15 & 3000 & 0.65 & 2 & 12 & 5 & 800 & 0.25 & 2 \\
Push-T($d=25$) & 5 & {10} & 10 & 20 & 900 & 0 & 2 & 10 & 30 & 300 & 0 & 5 \\
Push-T($d=50$) & 5 & {20} & 10 & 40 & 1500 & 0.9 & 4 & 10 & 20 & 900 & 0.8 & 5 \\
Push-T($d=75$) & 5 & {25} & 10 & 20 & 1200 & 0.9 & 5 & 10 & 20 & 1200 & 0.8 & 5 \\
\bottomrule
\end{tabular}
\par\vspace{10pt}
\setlength{\tabcolsep}{5pt}
\renewcommand{\arraystretch}{1.15}

\captionsetup{position=top,skip=5pt}
\captionof{table}{{Single-level MPPI planning hyperparameters for Diverse Maze.}}
\label{tab:mppi_l1}
\begin{tabular}{lcccccc}
\toprule
 & $k$ & {$B$} & noise $\sigma$ & \# samples & $\lambda$ & $h$ \\
\midrule
Maze ($D \in [5,8]$) & 4 & {500} & 5 & 125 & 0.0025 & 150 \\
Maze ($D \in [9,12]$) & 4 & {500} & 5 & 250 & 0.0025 & 200 \\
Maze ($D \in [13,16]$) & 4 & {500} & 5 & 250 & 0.0025 & 250 \\
\bottomrule
\end{tabular}
\par\vspace{10pt}

\captionsetup{position=top,skip=5pt}
\captionof{table}{{Hierarchical MPPI planning hyperparameters for Diverse Maze.}}
\label{tab:mppi_l2}
\begin{tabular}{lcccccccccc}
\toprule
 & & & \multicolumn{4}{c}{High-level planner} & \multicolumn{4}{c}{Low-level planner} \\
\cmidrule(lr){4-7}\cmidrule(lr){8-11}
 & $k$ & {$B$} & noise $\sigma$ & \# samples & $\lambda$ & $H$
 & noise $\sigma$ & \# samples & $\lambda$ & $h$ \\
\midrule
Maze ($D \in [5,8]$) & 4 & {265} & 10 & 2000 & 0.0025 & 25 & 5 & 500 & 0.0025 & {10} \\
Maze ($D \in [9,12]$) & 4 & {365} & 10 & 2000 & 0.0025 & 35 & 5 & 500 & 0.0025 & {10} \\
Maze ($D \in [13,16]$) & 4 & {485} & 10 & 4000 & 0.0025 & 47 & 5 & 1000 & 0.0025 & {10} \\
\bottomrule
\end{tabular}
\vspace{-1em}
\end{table}

For Diverse Maze, we use Model Predictive Path Integral (MPPI) \citep{williams2017information} for planning. MPPI updates a nominal action sequence by importance-weighting sampled trajectories using exponential costs; unlike CEM, it applies soft weights over all samples, yielding smoother updates without hard elite selection. We use the MPPI implementation from PLDM \citep{sobal2025learning}. See \cref{tab:mppi_l1} and \cref{tab:mppi_l2} for the single-level and hierarchical planner hyperparameters, respectively.

{In Diverse Maze's hierarchical planning setup, the hierarchical planner is called at each replanning step during the first $hH$ steps. For the final 15 steps, we use only the low-level planner, which plans directly toward the goal with a fixed prediction horizon of 50.}

\section{Computational Analysis Details}
\label{app:comp_study}

\subsection{Push-T}

\begin{figure}[h]
  \centering
  \includegraphics[width=0.8\columnwidth]{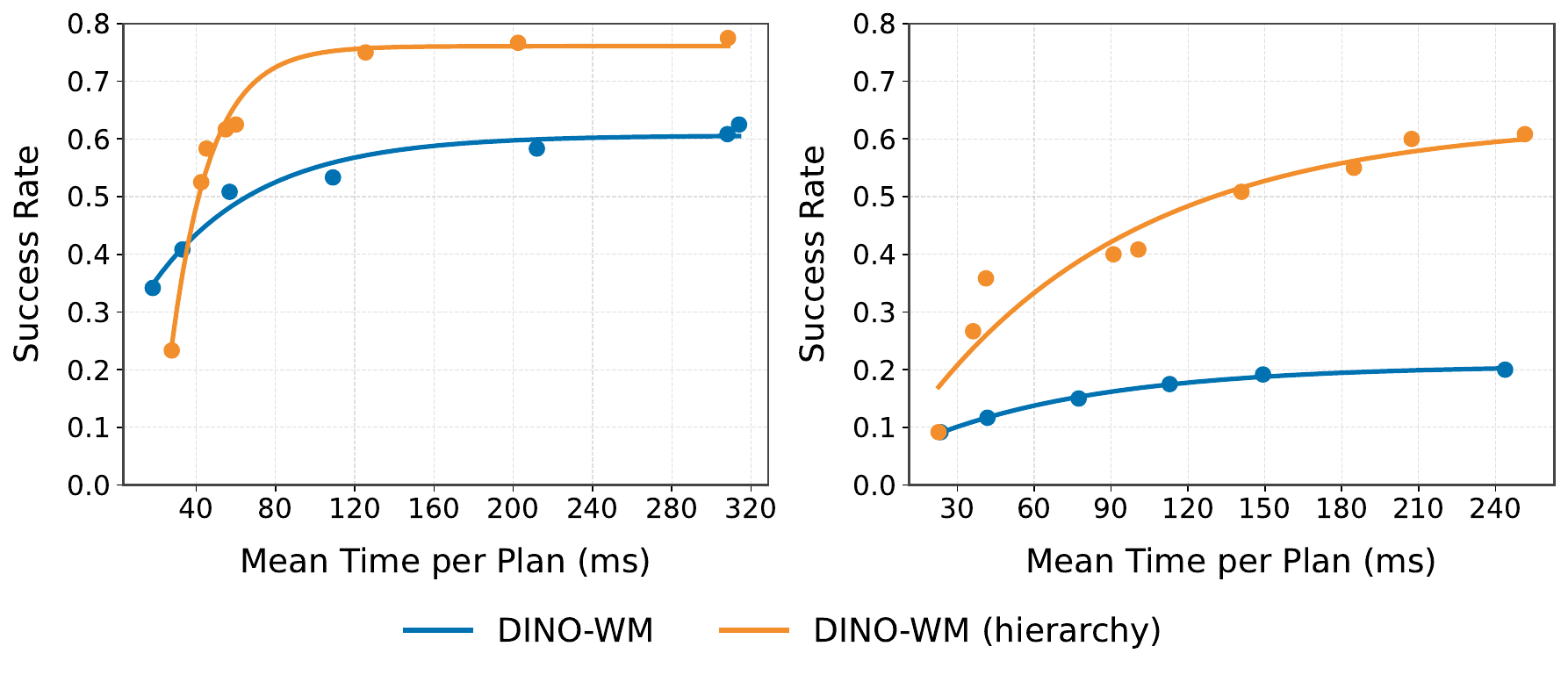}
  \vspace{-1em}
  \caption{Success rate as a function of test-time compute for Push-T. \textbf{Left}: Setting $d = 50$. \textbf{Right}: Setting $d = 75$.}
  \label{fig:pusht_compute_full}
\vspace{-1em}
\end{figure}

For the compute scaling analysis of \cref{fig:compute_analysis} and \cref{fig:pusht_compute_full}, we performed a grid search over the key CEM hyperparameters of both high-level and low-level planners. 

For $d=50$ in \cref{fig:compute_analysis}, for the high level planner, we swept $\#\text{samples} \in [150, 300, 600, 900, 1200, 1500]$, $\#\text{iters} \in [10, 20, 30, 40]$, and $\text{Var EMA} \in [0.4, 0.7, 0.9]$. For the low-level planner (both when used standalone and as the low-level planner in the hierarchical setting), we swept $\#\text{samples} \in [150, 300, 600, 900, 1200]$, $\#\text{iters} \in [10, 20, 30, 40]$, and $\text{Var EMA} \in [0, 0.4, 0.7, 0.8, 0.9]$.

For $d=75$, for the high-level planner, we swept the number of samples $\#\text{samples} \in [150, 300, 600, 900, 1200, 1500]$, the number of CEM iterations $\#\text{iters} \in [10, 20, 30, 40]$, and the standard deviation momentum coefficient $\text{Var EMA} \in [0.4, 0.7, 0.9]$. For the low-level planner, we swept $\#\text{samples} \in [150, 300, 600, 900, 1000, 1100, 1200, 1500]$, $\#\text{iters} \in [10, 20, 25, 30, 35, 40]$, and $\text{Var EMA} \in [0, 0.4, 0.5, 0.7, 0.8, 0.9, 0.95]$. 

This sweep over CEM hyperparameters enables constructing Pareto frontiers that characterize the trade-off between planning compute (mean time per plan) and task success rate.

\subsection{Diverse Maze}

\begin{figure}[h]
  \centering
  \includegraphics[width=0.8\columnwidth]{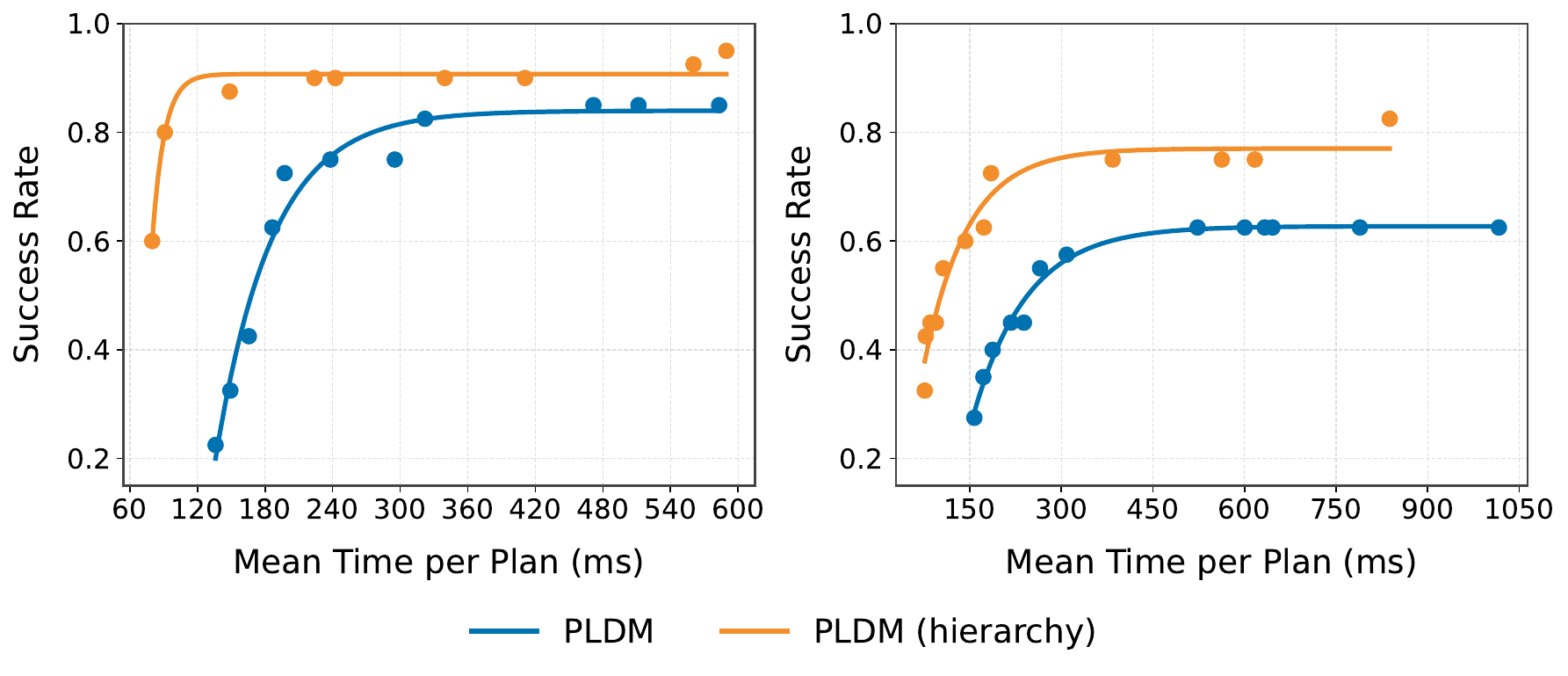}
  \vspace{-1em}
  \caption{Success rate as a function of test-time compute for Diverse Maze. \textbf{Left}: Setting $D \in [9,12]$. \textbf{Right}: Setting $D \in [13,16]$.}
  \label{fig:maze_compute_full}
\vspace{-1em}
\end{figure}

\label{app:comp_study_maze}

For the compute scaling analysis of the Diverse Maze environment in \cref{fig:compute_analysis} and \ref{fig:maze_compute_full}, we performed a grid search over the number of samples and horizon of planning in the MPPI algorithm. We perform the same sweep for all 3 difficulty settings.

For the flat baseline, we swept the number of samples $\#\text{samples} \in [70, 125, 250, 500, 1000, 2000, 4000]$ and the horizon $ \in [100, 125, 150, 175, 200, 225, 250, 275]$. 

For the hierarchical planner, we have a low-level planner and a high-level planner, each of which has a number of samples. We swept the number of samples of the low-level planner $\#\text{samples} \in [70, 125, 250, 500, 1000, 2000, 4000]$, the number of samples of the high-level planner $\#\text{samples} \in [70, 125, 250, 500, 1000, 2000, 4000]$ and the horizon of the high-level planner $H \in [25,35,45]$. This comprehensive sweep over MPPI hyperparameters enables the construction of Pareto frontiers that characterize the trade-off between planning compute time and the success rate.

\section{Additional Ablations and Analysis}

\subsection{Model Size vs. Hierarchy}
\label{app:capacity}

One question is whether the gains from hierarchical planning arise from compositional structure at planning time, or simply from increased model capacity. To disentangle these effects, we compare hierarchical models against flat world models with matched or larger parameter counts. As shown in Table~\ref{tab:ablation-sizes}, increasing the capacity of flat models does not improve performance—in fact, it often degrades it—while hierarchical models consistently achieve higher success rates across both tasks. This suggests that the gains stem from hierarchical composition rather than model size alone.

\begin{table}[h]
\small
\centering
\renewcommand{\arraystretch}{1.2}
\setlength{\tabcolsep}{4pt}
\caption{Capacity-controlled comparison between flat and hierarchical world models. Flat models with comparable or larger parameter counts fail to match hierarchical performance, and sometimes degrade with increased capacity. In contrast, hierarchical models consistently outperform flat baselines on Push-T and Diverse Maze, indicating that gains arise from hierarchical structure rather than model capacity.}
\label{tab:ablation-sizes}
\begin{minipage}{0.48\linewidth}
\centering
\subcaption{Push-T}
\label{tab:hierarchy-ablation-pusht}
\begin{tabular*}{\linewidth}{@{\extracolsep{\fill}} l @{\hspace{3pt}} c c c @{}}
\toprule
Method & Params & $d{=}50$ & $d{=}75$ \\
\midrule
DINO-WM (flat)         & $44$M & $55\%$ & $17\%$ \\
DINO-WM (flat) & $98$M & $35\%$ & $15\%$ \\
\midrule
DINO-WM (hierarchy)    & $94$M & \textbf{78\%} & \textbf{61\%} \\
\bottomrule
\end{tabular*}

\end{minipage}
\hfill
\begin{minipage}{0.48\linewidth}
\centering
\subcaption{Diverse Maze}
\label{tab:hierarchy-ablation-maze}
\begin{tabular*}{\linewidth}{@{\extracolsep{\fill}} l @{\hspace{3pt}} c c c @{}}
\toprule
Method & Params & $D\!\in\![9,12]$ & $D\!\in\![13,16]$ \\
\midrule
PLDM (flat)         & $54$k  & $85\%$ & $63\%$ \\
PLDM (flat) & $178$k & $82\%$ & $59\%$ \\
\midrule
PLDM (hierarchy)    & $182$k & \textbf{95\%} & \textbf{83\%} \\
\bottomrule
\end{tabular*}

\end{minipage}

\vspace{-1em}
\end{table}

\subsection{Strides for Fixed Interval Waypoints}

For Diverse Maze, the high-level world model is trained using fixed temporal strides between successive latent waypoints. We evaluate the sensitivity of hierarchical planning performance to this stride by varying it across a range of values. As shown in Table~\ref{tab:stride-ablation}, performance improves as the stride increases from 6 to 12 timesteps, suggesting that moderate temporal abstraction is beneficial for long-horizon planning. However, performance slightly degrades at larger strides, indicating a trade-off between abstraction and predictive fidelity. Overall, the results demonstrate that hierarchical planning is robust to the choice of stride, with strong performance across a broad range of values.

\begin{table}[h]
\small
\centering
\renewcommand{\arraystretch}{1.2}
\setlength{\tabcolsep}{6pt}
\caption{Sensitivity of hierarchical planning performance to the high-level temporal stride on Diverse Maze.}
\label{tab:stride-ablation}
\begin{tabular}{lccccc}
\toprule
High-level stride & 6 & 8 & 10 & 12 & 14 \\
\midrule
Success Rate ($D\!\in\![13,16]$)  & 50\% & 55\% & 77\% & 78\% & 75\% \\
\bottomrule
\end{tabular}

\vspace{-1em}
\end{table}

\subsection{Emergence of semantic subgoals}
\label{app:subgoals}

\begin{figure}[h]
\vspace{-2em}
  \centering
  \begin{minipage}[t]{0.49\linewidth}
    \centering
    \includegraphics[width=\linewidth]{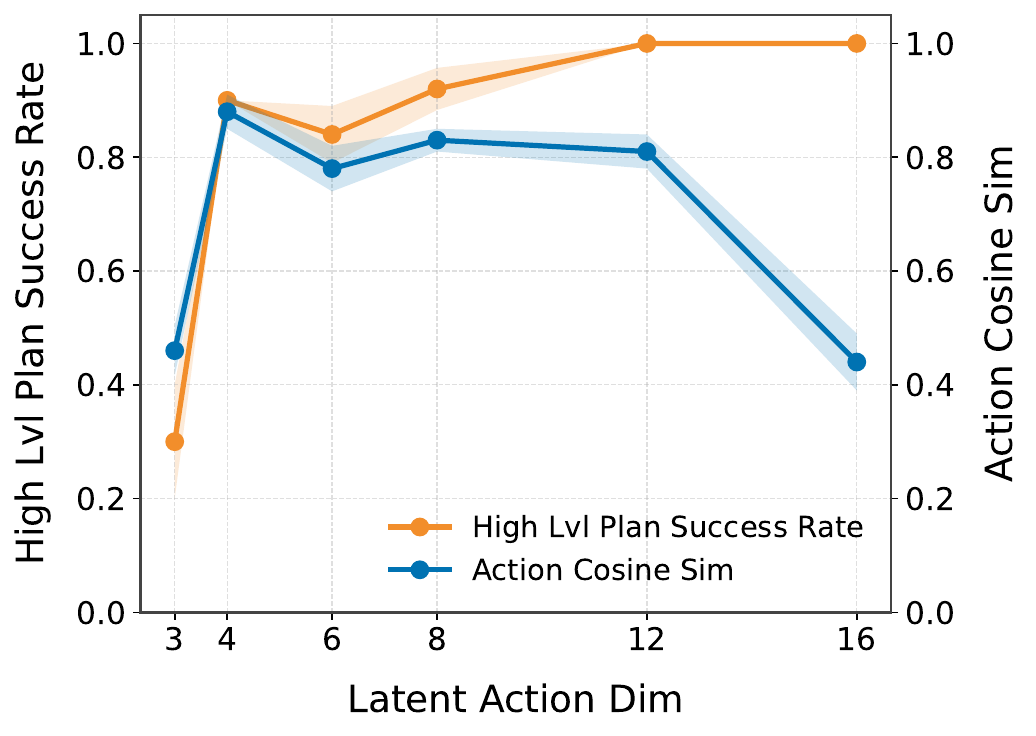}
  \end{minipage}\hfill
  \begin{minipage}[t]{0.49\linewidth}
    \centering
    \includegraphics[width=\linewidth]{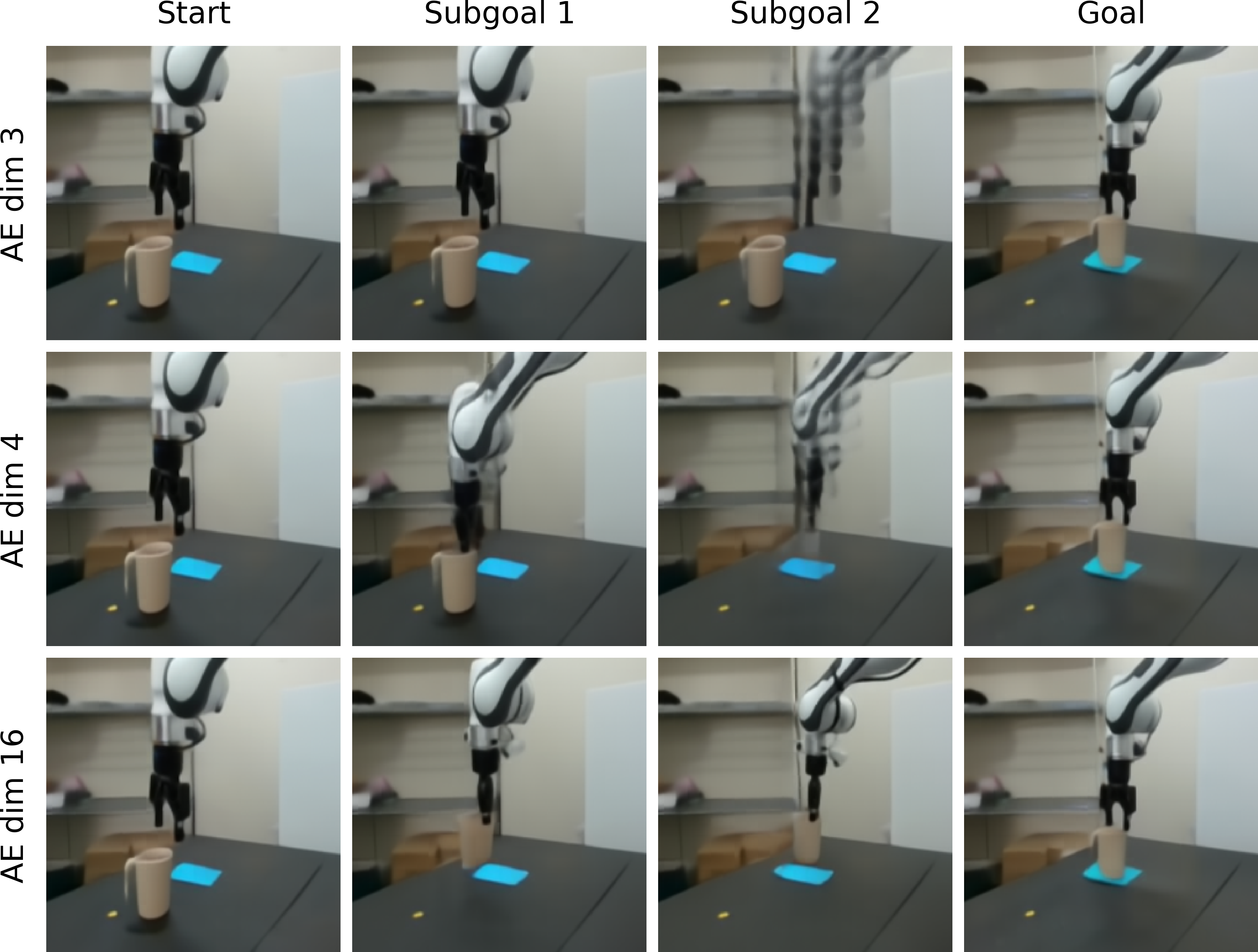}
  \end{minipage}

  \caption{
\textbf{Effect of macro-action dimension on hierarchical planning.}
\textbf{Left:} Performance as a function of macro-action dimension $d$.
When the latent space has very low capacity, the high-level planner fails to produce valid plans.
As $d$ increases, success improves; however, beyond a moderate dimensionality, proposed subgoals become harder for the low-level planner to execute, as indicated by a drop in action cosine similarity to expert behavior.
\textbf{Right:} Qualitative rollouts illustrating this trade-off.
Moderate latent dimensionality biases the planner toward reachable, greedy subgoals, yielding the best overall performance.
  }
  \label{fig:latent_dim_ablation}
  \vspace{-1em}
\end{figure}

Effective hierarchical planning requires subgoals on a feasible path to the final goal that are reachable by the low-level planner. High-level plans therefore fall into three categories: invalid plans, valid plans with unreachable subgoals, and valid plans with reachable subgoals.

As shown in \cref{fig:latent_dim_ablation}, when the macro-action space has sufficient capacity ($\geq 4$ dimensions), the high-level planner typically produces valid plans. However, these subgoals are not always reachable by the low-level planner, as they may require non-greedy action sequences. Restricting the latent-action dimensionality biases the planner toward proposing subgoals that are achievable with greedy behavior. This suggests an optimal regime in which the latent space is expressive enough to encode useful trajectories, but not so expressive that it enables subgoals requiring complex non-greedy execution. Empirically, a macro-action dimension of 4 strikes this balance for the Franka tasks.

Notably, reconstruction fidelity is not tightly correlated with latent-action dimensionality. Lower-dimensional latent spaces yield noisier predictions, reflected in higher $\ell_1$ error and blurrier reconstructions (\cref{fig:latent_dim_ablation}), as expected under stronger compression. Nevertheless, these predictions often preserve coarse semantic structure, such as contact events or motion direction, which is sufficient for hierarchical planning despite reduced visual precision.

\section{Manual Subgoals: Isolating Planning from Representation}
\label{app:manual_subgoals}

The flat VJEPA2-AC planner fails entirely on end-to-end pick-\&-place (Table~\ref{tab:franka}), but this aggregate failure conflates two possible causes: the flat planner may lack the {ability to accurately predict the consequences of the required actions}, or single-level MPC may be unable to escape the greedy-shortcut traps under a non-greedy goal matching objective (Section~\ref{sec:intro}). This appendix disentangles the two by supplying the flat planner with manually-defined intermediate subgoals.

\textbf{Setup.} For each of the 10 pick-\&-place configurations per object, we manually prepare an intermediate grasp-point subgoal image. The flat VJEPA2-AC planner is run in two sequential phases: first reaching the subgoal, then reaching the final goal. Octo is evaluated under the same image-subgoal supervision; $\pi_0$-FAST-DROID and $\pi_{0.5}$-DROID are conditioned on language goals and cannot consume image subgoals in the same way, so we omit them from this comparison.

\begingroup
\setlength{\intextsep}{0pt}
\begin{wraptable}{r}{0.48\columnwidth}
\centering
\small
\caption{{\textbf{Manual subgoals on Franka.} Success rates (\%) over 50 trials per object (10 configurations, 5 per configuration).}}
\label{tab:franka_full}
{%
\begin{tabular}{@{}llcc@{}}
\toprule
Planner & Subgoal & Cup & Box \\
\midrule
VJEPA2-AC & None & 0\% & 0\% \\
HWM & Predicted & 70\% & 60\% \\
\midrule
VJEPA2-AC & Manual & 80\% & 80\% \\
HWM & Manual & 80\% & 80\% \\
Octo & Manual & 20\% & 10\% \\
\bottomrule
\end{tabular}}
\end{wraptable}

\textbf{Results.} Table~\ref{tab:franka_full} reports pick-\&-place results with and without manual subgoals. With manual subgoals, the flat VJEPA2-AC planner reaches 80\%/80\% on pick-\&-place cup/box — comparable to HWM under the same supervision (also 80\%/80\%) and far above its 0\%/0\% without subgoals. The VJEPA2-AC world model is therefore competent on these tasks given the right intermediate targets; the failure of single-level planning is hierarchical in nature, not representational. HWM recovers most of this oracle performance automatically (70\%/60\% from a single goal image), confirming that its contribution is to produce reachable subgoals without manual annotation. Octo, despite receiving the same subgoals, achieves only 20\%/10\%, indicating that subgoal supervision alone does not close the gap — the underlying dynamics model must also support precise control.\par
\endgroup

\section{Visualizing World Model Predictions}

\textbf{Franka Arm with Robotiq Gripper.}
\label{app:franka_decoder}
To reconstruct RGB images from latent representations, we train on DROID a ViT decoder that maps the frozen V-JEPA 2 encoder features back to pixel space. The decoder takes as input the encoded features of dimension 1408 (matching the V-JEPA 2 ViT-Giant encoder output) and projects them to an intermediate dimension of 1024 via a linear layer. The architecture consists of 12 transformer blocks, each with 32 attention heads and an MLP ratio of 4, using GELU activations, LayerNorm and sinusoidal 2D positional embeddings. The decoder operates on a $32\times 32$ grid of spatial tokens (corresponding to patch size 8 for $256\times 256$ images). The model is trained using an $L_1$ reconstruction loss on normalized pixel values. We train the decoder for 100 epochs on the DROID dataset using the AdamW optimizer with a peak learning rate of $5\times 10^{-4}$, linear warmup over 2 epochs, cosine decay to $10^{-6}$, and weight decay of $0.1$. Training is performed on 8 nodes with a total batch size of 512.

\textbf{Push-T.}
Similarly as for the DROID decoder depicted above, we train a ViT decoder that maps the frozen DINOv2 ViT-S/14 encoder features back to pixel space on the Push-T dataset. The decoder is a ViT-Base of intermediate dimension of 768 with 12 transformer blocks, each with 16 attention heads. It operates on a $28\times 28$ grid of spatial tokens (corresponding to patch size 8 for $224\times 224$ images). The model is trained using an $L_2$ reconstruction loss on normalized pixel values for 40 epochs on the PushT dataset with a total batch size of 256.

\begin{figure}[h]
    \centering
    \includegraphics[width=\linewidth]{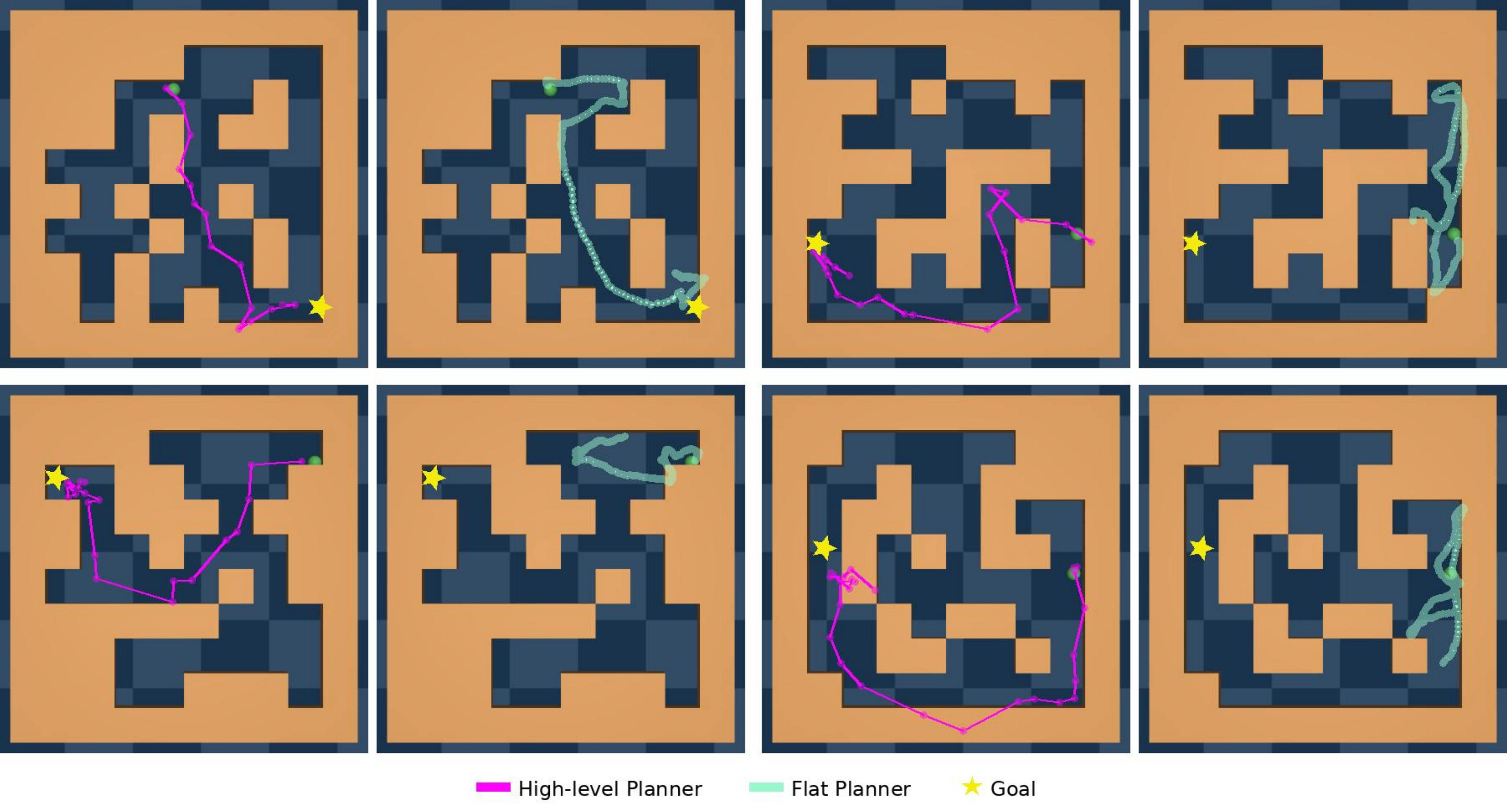}
    \caption{\textbf{High-level plans in Diverse Maze.} The agent is shown as a green ball and the goal as a yellow star. The pink trajectory is generated by the high-level planner, while the teal trajectory is generated by the flat planner. In these trials, the high-level planner successfully produces a sequence of subgoals that lead to the goal, whereas the flat planner often struggles due to the increased search complexity of long-horizon prediction.}

    \label{fig:maze_plans}
\end{figure}

\textbf{Diverse Maze.} To visualize predictions and plans, we train a lightweight convolutional prober that takes as input a spatial representation of shape $(H, W, C)$. The prober consists of two convolution–pooling stages followed by a fully connected layer, and outputs 2-D coordinates corresponding to the agent’s $(x, y)$ location. We show visualizations of successful high-level plans in \cref{fig:maze_plans}.

\clearpage
\section{Robot Task Definitions}
\label{app:robot_tasks}

\begin{figure}[h]
    \centering
    \includegraphics[width=\linewidth]{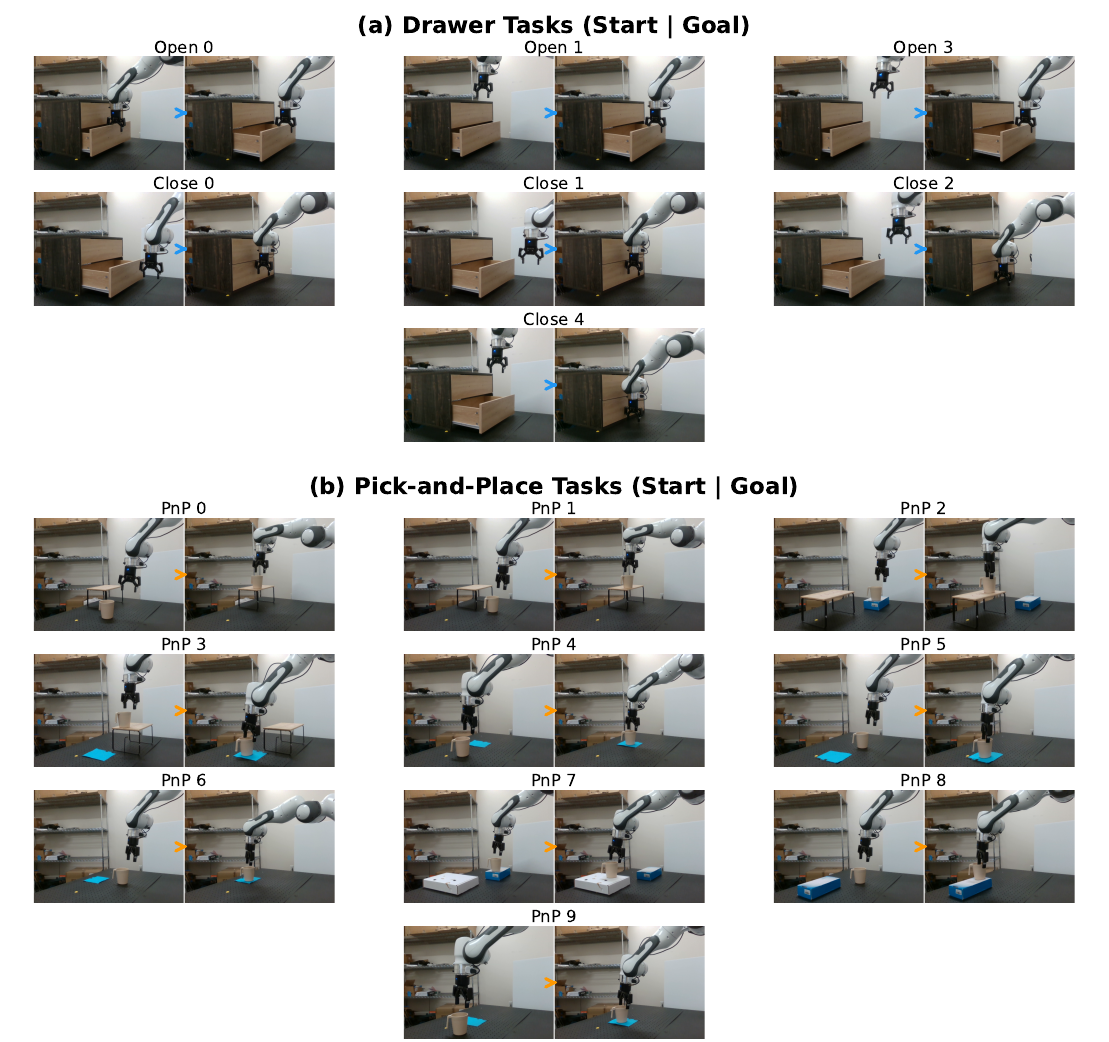}
    \caption{Initial and goal states for all evaluation tasks. \textbf{(a) Drawer Tasks:} 7 subtasks involving opening and closing drawers at various positions. \textbf{(b) Pick-\&-Place Cup Tasks:} 10 subtasks requiring the robot to pick up a cup and place it at different target locations. Each panel shows the initial state (left) and goal state (right) for a given task.}
    \label{fig:init_goal_states}
\end{figure}

\cref{fig:init_goal_states} depicts the initial and goal states defining the evaluation episodes for the Pick-\&-Place Cup and Drawer tasks. We omit the Pick-\&-Place Box tasks, which use identical start and goal configurations and differ only in the grasped object (a box rather than a cup).

\clearpage

\section{Per-Configuration Trial Breakdown for Franka}
\label{app:per-config}

We run $N=5$ independent trials per (start, goal) configuration for both
VJEPA2-AC and HWM on all pick-\&-place and drawer tasks, yielding $50$ total
trials per object on pick-\&-place and $35$ on drawer. Aggregate success rates
are reported in Table~\ref{tab:franka}; per-configuration breakdowns and
95\% Clopper--Pearson confidence intervals are in
Table~\ref{tab:franka_per_config}.

Per-configuration outcomes are highly consistent. For VJEPA2-AC, every
configuration is fully deterministic (5/5 identical outcomes across all $27$
configurations, $135/135$ trials). For HWM, $21/27$ configurations are fully
deterministic, and the remaining $6$ reproduce the modal outcome in $4/5$
trials ($129/135 = 95.6\%$ of trials match modal). Aggregate success rates
therefore reflect per-configuration task difficulty rather than execution
stochasticity.

\begin{table}[h]
\centering
\caption{Per-configuration success counts (successes / trials) on Franka,
with 95\% Clopper--Pearson confidence intervals on aggregate rates.
Configurations correspond to the start--goal pairs visualized in
\cref{app:robot_tasks}.}
\label{tab:franka_per_config}
\vspace{0.5em}

\begin{subtable}{0.48\linewidth}
\centering
\small
\caption{Pick-\&-Place Cup}
\begin{tabular}{lcc}
\toprule
Config & VJEPA2-AC & HWM (ours) \\
\midrule
PnP 0 & 0/5 & 0/5 \\
PnP 1 & 0/5 & 5/5 \\
PnP 2 & 0/5 & 5/5 \\
PnP 3 & 0/5 & 5/5 \\
PnP 4 & 0/5 & 5/5 \\
PnP 5 & 0/5 & 4/5 \\
PnP 6 & 0/5 & 1/5 \\
PnP 7 & 0/5 & 5/5 \\
PnP 8 & 0/5 & 0/5 \\
PnP 9 & 0/5 & 5/5 \\
\midrule
Total   & 0/50 & 35/50 \\
95\% CI & {\scriptsize [0.0, 7.1]} & {\scriptsize [55.4, 82.1]} \\
\bottomrule
\end{tabular}

\end{subtable}
\hfill
\begin{subtable}{0.48\linewidth}
\centering
\small
\caption{Pick-\&-Place Box}
\begin{tabular}{lcc}
\toprule
Config & VJEPA2-AC & HWM (ours) \\
\midrule
PnP 0 & 0/5 & 0/5 \\
PnP 1 & 0/5 & 1/5 \\
PnP 2 & 0/5 & 4/5 \\
PnP 3 & 0/5 & 5/5 \\
PnP 4 & 0/5 & 5/5 \\
PnP 5 & 0/5 & 4/5 \\
PnP 6 & 0/5 & 1/5 \\
PnP 7 & 0/5 & 5/5 \\
PnP 8 & 0/5 & 0/5 \\
PnP 9 & 0/5 & 5/5 \\
\midrule
Total   & 0/50 & 30/50 \\
95\% CI & {\scriptsize [0.0, 7.1]} & {\scriptsize [45.2, 73.6]} \\
\bottomrule
\end{tabular}

\end{subtable}

\vspace{1em}

\begin{subtable}{\linewidth}
\centering
\small
\caption{Drawer (opening and closing)}
\begin{tabular}{lccccccc|cc}
\toprule
Config & Open 0 & Open 1 & Open 3 & Close 0 & Close 1 & Close 2 & Close 4 & Total & 95\% CI \\
\midrule
VJEPA2-AC  & 5/5 & 0/5 & 0/5 & 5/5 & 0/5 & 0/5 & 0/5 & 10/35 & {\scriptsize [14.6, 46.3]} \\
HWM (ours) & 5/5 & 5/5 & 0/5 & 5/5 & 5/5 & 5/5 & 0/5 & 25/35 & {\scriptsize [53.7, 85.4]} \\
\bottomrule
\end{tabular}

\end{subtable}

\end{table}

\clearpage

\section{Additional Execution Rollouts for Franka}
\label{app:franka_rollouts}
\begin{figure}[h]
    \centering
    \includegraphics[width=\linewidth]{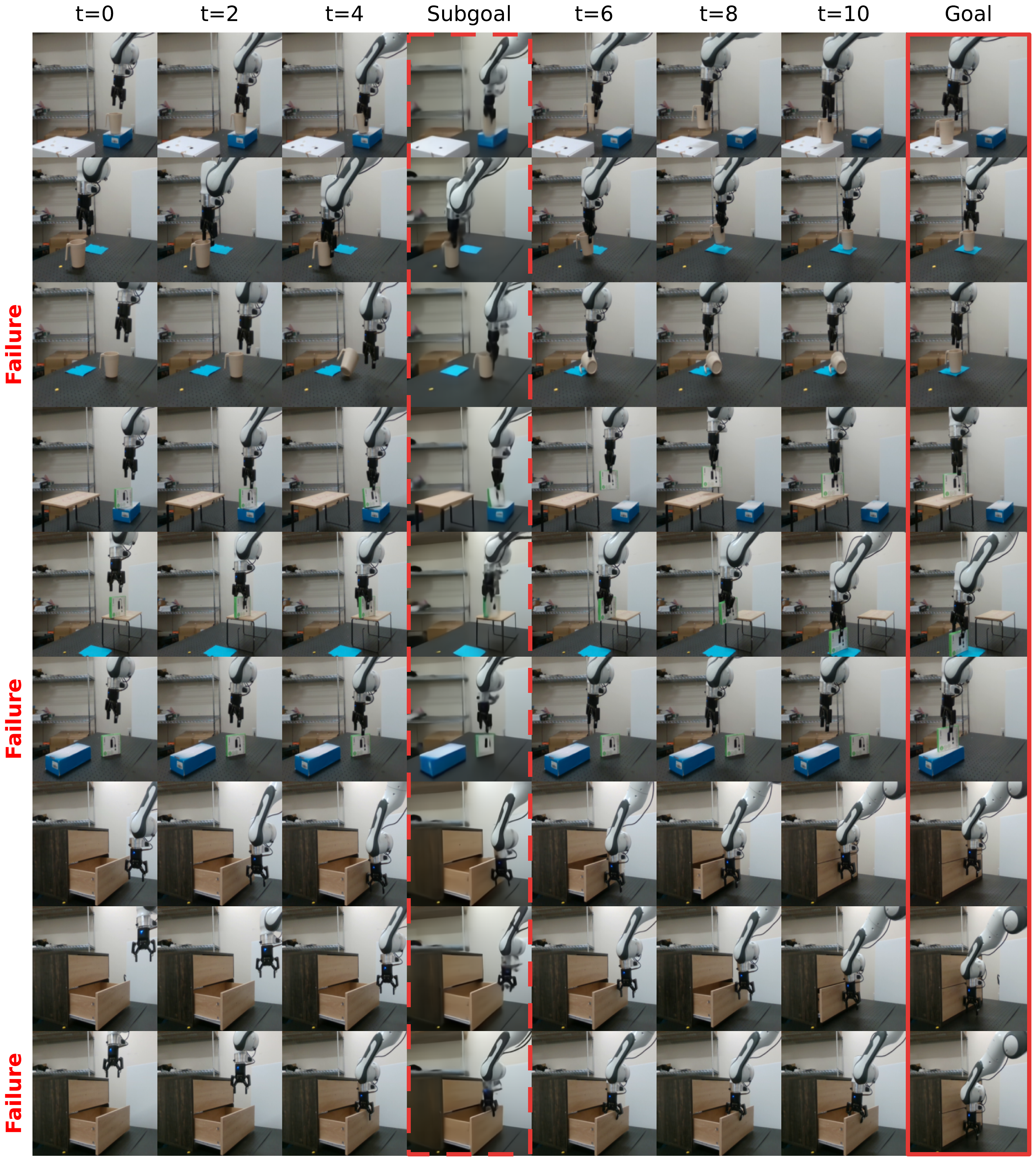}
    \caption{Executions on pick-\&-place and drawer manipulation with a dual-gripper Franka arm. Dotted red columns indicate subgoals inferred by the high-level planner and decoded for visualization. Rows representing failed episodes are labeled with "Failure". Additional execution videos can be found in \url{https://kevinghst.github.io/HWM/}.}
    \vspace{-1em}
\end{figure}

\clearpage

\section{Execution Rollout Examples for Push-T}

\begin{figure*}[h]
\vspace{-1mm}
  \centering
  \includegraphics[width=\textwidth]{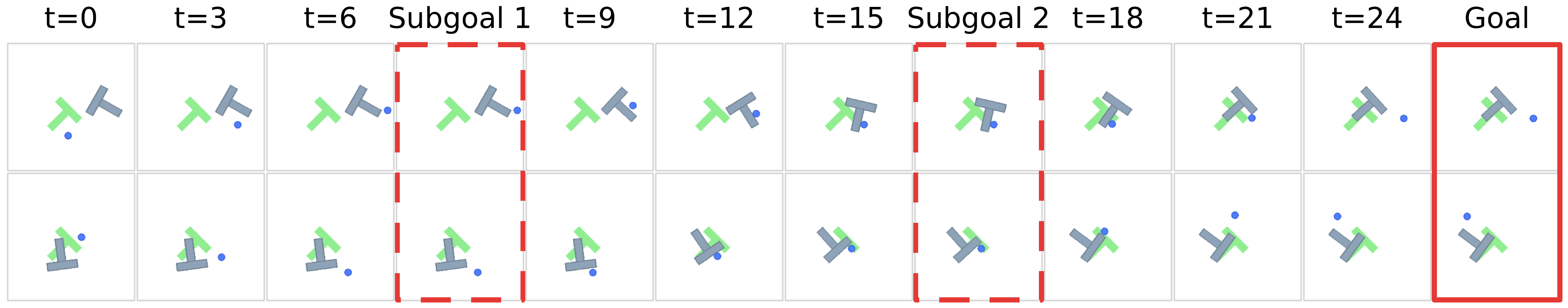}
  \caption{Executions on Push-T task. Dotted red columns indicate subgoals inferred by the high-level planner and decoded for visualization.}
  \label{fig:pusht_execution}
\vspace{-3mm}
\end{figure*}

\section{Offline Planning Evaluation for Franka}
\label{app:offline_plan_metric}
Since planning on a real robot is time-consuming and not parallelizable, we introduce a quantitative offline planning evaluation that enables faster iteration over model variants and planning hyperparameters. We evaluate on the same 10 pick-\&-place tasks used in the interactive experiments, with manually defined subgoals corresponding to the grasp point. Expert actions are defined as the robot’s delta pose from the initial observation to the subgoal, $\tilde{a}_1$, and from the subgoal to the goal, $\tilde{a}_2$.
Conditioned on the initial and goal observations, the hierarchical planner produces a sequence of subgoals along with a sequence of primitive actions $(a_{1:h}^*)$ toward the first subgoal. To assess high-level planning quality, we compare the aggregate inferred action $\sum_{i=1}^{h} a^*_i$ with the expert action $\tilde{a}_1$ using cosine similarity and $\ell_1$ distance. High cosine similarity or low $\ell_1$ distance indicates that the planner has identified a plan that moves toward the non-greedy subgoal, which strongly correlates with real-world task success.

\section{Baselines}
\label{app:baselines}

\subsection{Vision-Language-Action Models}
\label{app:vla}

For our baselines, we reproduce the same setup as in the Droid platform. Specifically, we use a Franka Panda arm with a bi-finger Robotiq gripper, dual camera setup (one wrist camera plus one side camera looking at the robot from the left). Our wrist camera setup is identical to Droid (the same mount and model - ZED Mini), however, for the side camera, we use Intel RealSense. While RealSense has a different aspect ratio (4:3 instead of ZED Mini's 16:9) we found that simply padding the side image to get 16:9 aspect ratio yields good success rate for VLAs trained on Droid with this aspect ratio. Note that we use the same RealSense side camera for all baselines and world model-based agents therefore, they all have access to the same amount of sensor signal from the side view. In fact, the $\pi$- family of models receive extra sensor information since they have access to wrist camera which is not used by Octo and world model-based agents.

\textbf{$\mathbf{\pi_{0}}$-FAST-DROID and $\mathbf{\pi_{0.5}}$-DROID} \citep{pertsch2025fast,intelligence2025pi_}. For a reliable control with the $\pi_x$-DROID models family, we reproduce the same controller on our Franka as in the Droid benchmark. Both models use one side camera and one wrist camera and output chunks of actions being joint velocities and target gripper positions. For both models, we execute the entire chunk of outputted actions before re-querying the policy ($15$ actions for $\pi_{0.5}$ and $8$ actions for $\pi_{0}$-FAST). We run both policies for $300$ total actions as we found this to be enough for a successfull task attempt with a big space for suboptimality. We do not finetune these policies, therefore, the evaluation is zero-shot and out-of-distribution since they are only finetuned on Droid so our scene is new to both of them.

Since $\pi$- models use text goal specification, we tuned text prompts to achieve the highest success rate. We arrived at the same conclusions as in the RoboArena \citep{atreya2025roboarena} benchmark, specifically, $\pi$- models do not understand colors and complex nouns for shapes (e.g. platform, shelf etc) and each prompt has to be written in a short and a simple way. Below are the prompts we tried (either ``cup'' or only ``box'' is present in the prompt):
\begin{enumerate}
    \item ``put the cup on the plane''
    \item ``put the cup (box) on the platform''
    \item ``put the cup (box) on the table'' -- the model was picking the object up and putting it on the the same position since the ``ground level'' surface is a table despite a small table was in front of it (See Figure \ref{fig:init_goal_states}).
    \item ``put the cup (box) on the yellow table'' -- same as 2.
    \item ``put the cup on the wooden table'' -- same as 2.
    \item ``put the cup on the shelf'' -- the policy tried to place the object on the shelf in the background.
    \item ``put the cup on the plate'' -- this prompt worked. Since it requires a plate, we started to use the one as the marker for the goal position for the object placement. \textit{We believe makes language goals a good substitution for image goals as the plate position unambiguously informs the model where the object should be placed and the plate is present in many training episodes.}
\end{enumerate}

We run $\pi_{0.5}$-DROID for $100$ trials for the pick-\&-place task with the cup and the pick-\&-place task with the box. We use the same 10 task configurations as the ones present in Figure \ref{fig:init_goal_states} to ensure fair comparison. This means we run 10 trials for each configuration which gives a low variance for the policy with highly stochastic outputs. Similarly, we run $100$ trials for the pick-\&-place task with the cup and $50$ trials for the pick-\&-place task with the box (i.e. 5 trials per configuration).

Similarly to the evaluation of the world models and policies, we use binary success rate for evaluations. In particular, we simply give the policy a score of 1 if the object was placed in the target position and zero otherwise. Unfortunately, we were unable to run drawer tasks with $\pi$- models because during our evaluation of the drawer task with Octo, the drawer was physically destroyed by the robot into a non-operating state. Note that because of language goals (which can only be primitive), $\pi$- models cannot run the task with subgoals.

\textbf{Octo} \citep{team2024octo}. We use the same setup for octo as in the V-JEPA 2 paper. Specifically, we use the Octo-base policy finetuned on Droid with image goals. We used the same scoring as for the $\pi$- models. Note that Octo is not able to work well without subgoals so we found it performance without subgoals to be 0.

\subsection{Goal-Conditioned \& Zero-Shot RL Methods}
\label{app:gcrl}

\begin{table}[h]
\centering
\footnotesize
\renewcommand{\arraystretch}{1.1}
\setlength{\tabcolsep}{4pt}
\caption{\textbf{Hyperparameters.} Hyperparameters used for GCIQL, HIQL, and HILP on the Push-T and Diverse Maze tasks.}
\label{app:gcrl_hyperparams}
\begin{tabular}{l|ccc|cccc|cc} 
\toprule
 & \multicolumn{3}{c|}{GCIQL}
 & \multicolumn{4}{c|}{HIQL}
 & \multicolumn{2}{c}{HILP} \\
\cmidrule(lr){2-4}\cmidrule(lr){5-8}\cmidrule(lr){9-10}
 & lr & expectile & rand goal $p$
 & lr & expectile & rand goal $p$ & alpha
 & lr & expectile \\
\midrule
Push-T($d=25$) & $3\mathrm{e}{-5}$
 & 0.6 & 0.0  & $3\mathrm{e}{-5}$
 & 0.9 & 0.0  & 3.0 & $3\mathrm{e}{-5}$ & 0.6 \\
Push-T($d=50$) & $3\mathrm{e}{-5}$ & 0.6 & 0.3 & $3\mathrm{e}{-5}$ & 0.9 & 0.3  & 1.0 & $3\mathrm{e}{-5}$ & 0.6 \\
Push-T($d=75$) & $3\mathrm{e}{-5}$ & 0.9 & 0.6 & $3\mathrm{e}{-5}$ & 0.9 & 0.3 & 1.0 & $3\mathrm{e}{-5}$ & 0.6 \\
\midrule
Maze($D \in [5,8]$)   & $3\mathrm{e}{-5}$  & 0.8 & 0.0 & $3\mathrm{e}{-5}$ & 0.6 & 0.3 & 1.0 & $3\mathrm{e}{-5}$ & 0.9 \\
Maze($D \in [9,12]$)  & $3\mathrm{e}{-5}$  & 0.6 & 0.3 & $3\mathrm{e}{-5}$ & 0.8 & 0.6 & 3.0 & $3\mathrm{e}{-5}$ & 0.9 \\
Maze($D \in [13,16]$) & $1\mathrm{e}{-4}$ & 0.6 & 0.0 & $3\mathrm{e}{-5}$ & 0.8 & 0.6 & 3.0 & $3\mathrm{e}{-5}$ & 0.9 \\
\bottomrule
\end{tabular}

\end{table}

\textbf{GCIQL} \citep{park2023hiql} -- goal-conditioned version of Implicit Q-Learning \citep{kostrikov2021offline}, a strong and widely-used method for offline RL. \\
\textbf{HIQL} \citep{park2023hiql} -- a hierarchical GCRL method which trains two policies: one to generate subgoals, and another one to reach the subgoals. Notably, both policies use the same value function. \\
\textbf{HILP} \citep{park2024foundation} -- a method that learns state representations from the offline data such that the distance in the
learned representation space is proportional to the number of steps between two states. A direction-conditioned policy is then learned to be
able to move along any specified direction in the latent space. \\

For GCIQL and HIQL, we use the implementations from the repository\footnote{\label{ogbench_fn}\href{https://github.com/seohongpark/ogbench}{https://github.com/seohongpark/ogbench}} of OGBench \citep{park2024ogbench}. Likewise, for HILP we use the official implementation\footnote{\label{hilp_fn}\href{https://github.com/seohongpark/HILP}{https://github.com/seohongpark/HILP}} from its authors. For Push-T and Diverse Maze, we tune hyperparameters independently for each difficulty setting and report results using the optimal configurations. The selected hyperparameters are reported in \cref{app:gcrl_hyperparams}; all other hyperparameters not listed in the table use the default values from the original repositories.

\end{document}